\documentclass[sn-basic, iicol]{sn-jnl}
\usepackage{graphicx}%
\usepackage{multirow}%
\usepackage{amsmath,amssymb,amsfonts}%
\usepackage{amsthm}%
\usepackage[title]{appendix}%
\usepackage[dvipsnames]{xcolor}%
\usepackage{textcomp}%
\usepackage{manyfoot}%
\usepackage{booktabs}%
\usepackage{algorithm}%
\usepackage{algorithmic}%
\usepackage{listings}%

\definecolor{colorbaseline}{RGB}{235,235,235}
\definecolor{colorpsp}{RGB}{224,238,238}
\definecolor{colorist}{RGB}{225, 240, 213}
\definecolor{colorA2M2}{RGB}{255,214,165}
\definecolor{colorA2M2lt}{RGB}{255,245,205}
\definecolor{colormtfan}{RGB}{242,216,137}
\definecolor{grey}{HTML}{808080}
\definecolor{colorR}{HTML}{3370BD}
\definecolor{colortau}{HTML}{4CAF50}
\definecolor{colortauorg}{HTML}{FF82AB}
\definecolor{colororange}{HTML}{658533}
\definecolor{revisioncolor}{HTML}{3370BD}

\usepackage{bookmark}
\usepackage{booktabs}
\usepackage{graphicx}
\usepackage{subcaption} 
\usepackage{makecell}
\usepackage{tabu}
\usepackage{mathtools}  
\usepackage{float}
\usepackage{colortbl}
\usepackage{arydshln} 
\usepackage{lmodern}
\usepackage{microtype}
\usepackage{epsfig}
\usepackage{graphicx}
\usepackage{paralist}
\usepackage{stfloats}
\usepackage{booktabs}
\usepackage{multicol}
\usepackage{multirow}
\usepackage{bm}
\usepackage{bbm}
\usepackage{diagbox}
\usepackage{color, colortbl}
\usepackage{xcolor}
\usepackage{pifont}
\newcommand{\tablestyle}[2]{\setlength{\tabcolsep}{#1}\renewcommand{\arraystretch}{#2}\centering\footnotesize}

\def\ours{HENet}
\def\ourspp{HENet++}

\newcommand{\figref}[1]{Figure~\ref{#1}}%
\newcommand{\tabref}[1]{Table~\ref{#1}}%
\newcommand{\secref}[1]{Section~\ref{#1}}
\renewcommand{\eqref}[1]{Eq.~(\ref{#1})}

\newcommand{\blue}[1]{{\color{blue}{#1}}}
\newcommand{\red}[1]{{\color{red}{#1}}}
\newcommand{\gr}{\rowcolor[gray]{.96}}
\newcommand{\he}{\rowcolor[RGB]{245,250,244}}
\newcommand{\hepp}{\rowcolor[RGB]{234,243,233}}

\usepackage{hyperref}

\hypersetup{
    colorlinks=true,
    allcolors=cyan,
}

\begin{document}



\title[Article Title]{\ourspp: Hybrid Encoding and Multi-task Learning for 3D Perception and Planning in Autonomous Driving}


\author[1,2]{\fnm{Zhongyu} \sur{Xia}}\email{xiazhongyu@pku.edu.cn}

\author[1]{\fnm{Zhiwei} \sur{Lin}}\email{zwlin@pku.edu.cn}

\author*[1,2]{\fnm{Yongtao} \sur{Wang}}\email{wyt@pku.edu.cn}

\author[3]{\fnm{Ming-Hsuan} \sur{Yang}}\email{mhyang@ucmerced.edu}

\affil[1]{\orgdiv{Wangxuan Institute of Computer Technology}, \orgname{Peking University}, \city{Beijing}, \country{China}}

\affil[2]{\orgname{VGI Labs Co., Ltd.}, \country{China}}

\affil[3]{\orgname{University of California, Merced}, \country{USA}}


\abstract{
Three-dimensional feature extraction and multi-task perception are fundamental components of modern autonomous driving systems. Although large image encoders, high-resolution inputs, and long temporal contexts can substantially improve representation quality and overall performance, jointly leveraging these strategies remains challenging due to prohibitive computational costs during both training and inference. Furthermore, different perception tasks often require distinct feature representations, making it difficult for a unified architecture to achieve end-to-end multi-task performance comparable to specialized single-task systems. To address these challenges, we propose HENet++, a unified framework for multi-task 3D perception and end-to-end autonomous driving that employs a hybrid image encoding strategy, using a large encoder for short-term frames and a lightweight encoder for long-term temporal context to strike a favorable balance between accuracy and efficiency. The framework jointly extracts dense background features and sparse foreground features, enabling task-specific representations that reduce cumulative errors and provide richer information for downstream prediction and planning modules. HENet++ is compatible with diverse 3D feature extraction pipelines and supports multi-modal inputs, including camera and radar data. Extensive experiments demonstrate state-of-the-art performance on the nuScenes multi-task 3D perception benchmark, achieving the lowest collision rate on the nuScenes planning benchmark and higher PDMS on the NAVSIM benchmark.
}

\keywords{Autonomous Driving, 3D Object Detection, BEV Segmentation, Occupancy Network, 3D Perception}

\maketitle


\section{Introduction}
\label{sec:introduction}

Three-dimensional perception constitutes a fundamental component of autonomous driving systems. Existing research in this area primarily focuses on designing effective sensor encoding strategies and decoder architectures to support diverse downstream perception tasks, including 3D object detection, Bird’s Eye View (BEV) semantic segmentation, and semantic occupancy prediction.
The encoding process extracts informative 3D feature representations from raw sensor data, and the multi-task decoding process utilizes these representations to generate predictions across multiple perception tasks.

In modern modular autonomous driving systems~\citep{mp3, occnet}, the outputs of 3D perception form the basis for subsequent planning and control. Nevertheless, the majority of the computational complexity in contemporary 3D perception models is caused by sensor data encoding.
This computational bottleneck has driven research toward multi-task learning frameworks, in which a shared encoder supports multiple perception tasks simultaneously, thereby improving computational efficiency.
More recently, end-to-end autonomous driving frameworks~\citep{UniAD, jiang2023vad, zheng2024genad} have been proposed to address these challenges by employing a unified network that directly outputs trajectory plans or control commands without relying on intermediate perception results. However, several studies~\citep{weng2024drive, zheng2024genad, sun2024sparsedrive} have shown that supervision based solely on trajectory-level signals is often insufficient to learn complex decision-making logic, leading to unstable behavior and hallucinated outputs. To alleviate these issues, recent end-to-end approaches incorporate explicit perception objectives to regularize 3D feature learning or constrain planning predictions. Consequently, despite differences in system design, both modular and end-to-end paradigms fundamentally depend on effective 3D feature encoding and robust multi-task 3D perception to ensure safety and reliability.

\begin{figure*}[t]
 \setlength{\abovecaptionskip}{10pt}
    \centering
    \includegraphics[width=.95\linewidth]{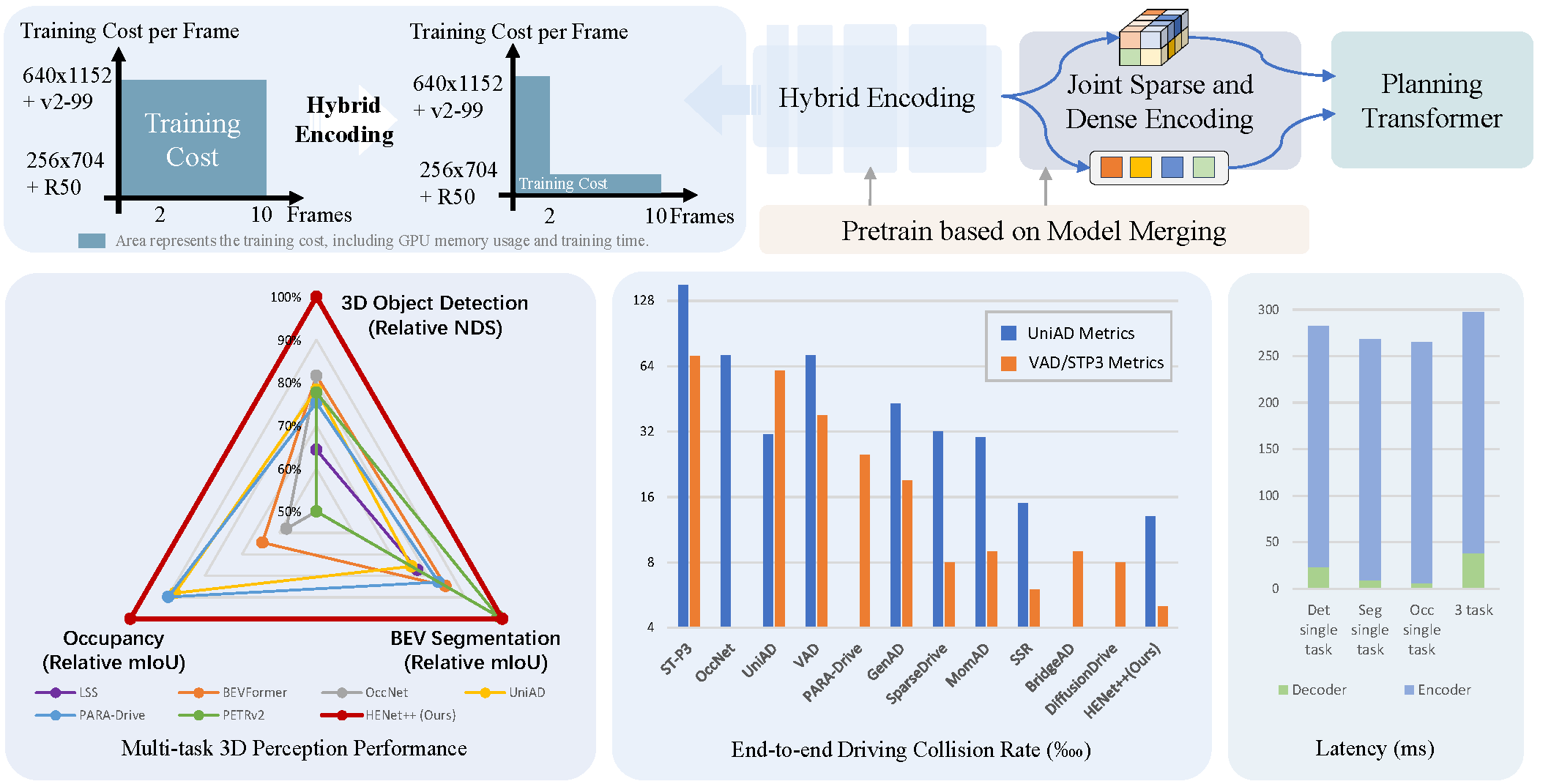}
\caption{
HENet++ reduces training cost by enabling the joint use of high-resolution images and long temporal sequences through hybrid encoding.
By integrating hybrid encoding, joint sparse and dense feature extraction, and pretraining via model merging, HENet++ achieves state-of-the-art multi-task performance while attaining the lowest collision rate in end-to-end driving on the nuScenes benchmark.
}
    \label{fig:intro}
\end{figure*}

Nonetheless, effective 3D feature encoding and multi-task perception remain challenging.
A primary difficulty lies in sensor information encoding, where simultaneously supporting high input resolution, large-capacity encoders, and long temporal contexts is nontrivial.
Autonomous driving requires reliable object recognition at distances of tens to hundreds of meters, motivating the use of high-resolution multi-view imagery.
Processing such inputs typically requires larger models with increased computational complexity across backbone networks, neck modules, and 2D-to-3D transformation components.
Furthermore, limited overlap among multi-camera views makes accurate spatial reasoning dependent on temporal cues, while occluded objects often require inference from historical frames.
However, jointly increasing resolution, model capacity, and temporal horizon significantly raises computational cost and training complexity.

The second challenge arises from task conflicts in multi-task learning. Different perception tasks often emphasize distinct aspects of the scene. For example, some focus primarily on foreground objects, whereas others concentrate on background structures, potentially favoring different architectural designs or feature representations. When the encoder is shared directly, the performance of individual tasks may degrade compared to specialized single-task models. Furthermore, incorporating pre-trained parameters originating from different task objectives can introduce additional optimization conflicts, affecting overall performance. How to effectively balance multiple tasks while improving joint performance remains a problem.


The third challenge concerns how to effectively leverage multi-modal information and advanced 3D perception models to build end-to-end autonomous driving systems. Existing end-to-end approaches adopt diverse intermediate subtasks, 3D representations, and planning module designs, resulting in complex system architectures without a clearly established unified paradigm.

To address the aforementioned challenges, we propose HENet~\citep{xia2024henet}, a preliminary framework based on multi-view camera inputs and dense BEV feature representations. 
To tackle the first challenge, we introduce a hybrid encoding strategy that employs a large neural network to encode a small number of high-resolution frames, while using a compact network to process long temporal sequences at lower resolution. This design enables the simultaneous utilization of high-resolution spatial information and long-term temporal context with relatively low computational overhead. 
To address the second challenge, we further propose a multi-task BEV feature encoding scheme tailored to BEV-based perception models.

Building on HENet, we propose HENet++, which improves compatibility and overall performance. 
To address the first challenge, we extend the hybrid encoding strategy to support not only dense feature extraction (e.g., BEV or voxel representations) but also sparse feature extraction, enabling compatibility with a broader range of existing 3D perception pipelines. 

For the second challenge, we analyze the representation preferences of different tasks and jointly extract sparse and dense features within a unified framework. Specifically, sparse features are used for tasks that emphasize foreground objects (e.g., 3D object detection), while dense voxel-based features are used for tasks that require structured spatial representations (e.g., BEV semantic segmentation or occupancy prediction). This design achieves strong multi-task perception performance on the nuScenes dataset. In addition, we introduce a model-merging-based pretraining strategy to further improve multi-task learning performance. 

To address the third challenge, we extend the HENet++ multi-task perception framework into an end-to-end autonomous driving system that predicts future object motions and generates trajectory plans for the ego vehicle. The proposed framework supports multimodal inputs from millimeter-wave radar and multi-view cameras, achieving a lower collision rate on the nuScenes benchmark compared to existing approaches. Besides, HENet++ achieves higher PDMS than existing end‑to‑end autonomous driving models on the NAVSIM benchmark.

The contributions of HENet++ are summarized as follows:

\begin{itemize}
\item [1)]
We extend Hybrid Encoding within the HENet++ framework to support both dense background features and sparse foreground features, enabling the joint utilization of long-term temporal context and high-resolution spatial information.

\item [2)]
We propose a multi-task learning framework that provides task-appropriate feature representations and enables end-to-end prediction for three perception tasks. In addition, we introduce a model-merging-based pretraining strategy to improve multi-task performance. HENet++ achieves state-of-the-art multi-task perception performance on the nuScenes dataset.

\item [3)]
Building upon the HENet++ perception framework, we develop an end-to-end autonomous driving model. The framework supports both vision‑only and radar‑camera multimodal inputs. On the nuScenes benchmark, HENet++ achieves a lower collision rate compared to existing approaches. Besides, HENet++ achieves higher PDMS than existing end‑to‑end autonomous driving models on the NAVSIM benchmark.
\end{itemize}


\section{Related Work}

\subsection{Multi-View 3D Object Detection}

3D object detection is a fundamental problem in 3D perception that aims to localize objects by predicting their 3D bounding boxes while estimating semantic categories and associated confidence scores.
Early approaches primarily relied on monocular imagery to infer 3D geometry and object localization~\citep{wang2019pseudo,Fcos3d,D4LCN,M3d-rpn,DFM,dd3d,CaDNN,OFT}.
With the rapid development of autonomous driving systems, multi-view camera setups have become standard sensing configurations, providing richer spatial and contextual cues for robust perception.
Existing multi-view 3D object detection approaches can be broadly divided into two paradigms based on view transformation strategies: BEV-based methods, which project image features into a unified bird’s-eye-view representation for spatial reasoning~\citep{BEVDet,BEVDepth,BEVDet4D,BEVStereo,SOLOFusion,aedet,fastBEV,polarformer,BEVFormer,bevformerv2,sts,HoP}, and sparse query-based methods, which directly model object instances in 3D space using learnable queries without explicitly constructing dense BEV representations~\citep{simMOD,DETR3D,PETR,PETRv2,3d-man,sparse4d,sparse4dv2,StreamPetr,SparseBEV,far3d}.

\vspace{1mm}
\noindent \textbf{BEV-based Methods}.
BEV-based approaches construct a unified bird’s-eye-view (BEV) representation from multi-view images to facilitate spatial reasoning.
BEVDet~\citep{BEVDet} adopts the Lift-Splat-Shoot (LSS)~\citep{LSS} framework with depth prediction, while BEVDepth~\citep{BEVDepth} improves depth estimation by incorporating camera parameters and LiDAR supervision.
Temporal extensions such as BEVDet4D~\citep{BEVDet4D}, BEVStereo~\citep{BEVStereo}, and STS~\citep{sts} leverage temporal alignment or stereo cues to enhance depth and motion modeling.
BEVFormer~\citep{BEVFormer} and Polarformer~\citep{polarformer} employ BEV queries with cross-attention for long-term temporal aggregation, further improved by BEVFormerv2~\citep{bevformerv2}.
Other works explore alternative temporal fusion or architectural designs, including hierarchical fusion (SOLOFusion~\citep{SOLOFusion}), historical object prediction (HoP~\citep{HoP}), and azimuth-equivariant modeling (AeDet~\citep{aedet}).

\vspace{1mm}
\noindent \textbf{Sparse Query-based Methods}.
Sparse query-based approaches directly model object instances using learnable queries without explicitly constructing dense BEV representations.
DETR3D~\citep{DETR3D} introduces sparse 3D object queries, later enhanced by PETR~\citep{PETR} and PETRv2~\citep{PETRv2} with positional and temporal embeddings.
Subsequent works focus on temporal modeling and efficiency, including memory-based aggregation (3D-MAN~\citep{3d-man}), multi-view and temporal keypoint projection (Sparse4D~\citep{sparse4d}, Sparse4Dv2~\citep{sparse4dv2}), efficient temporal representations (StreamPetr~\citep{StreamPetr}), perspective-aware aggregation (Far3D~\citep{far3d}), and adaptive spatio-temporal attention mechanisms (SparseBEV~\citep{SparseBEV}).

3D object detection primarily focuses on foreground objects, where sparse query-based methods exhibit distinct advantages over BEV-based approaches.
This advantage is largely attributed to the inherently sparse nature of object detection.
Even when dense features are constructed, object-centric representations must ultimately be extracted, introducing additional processing stages such as depth estimation, feature projection, and heatmap clustering, which may accumulate errors.
By directly modeling object instances, sparse query-based methods reduce intermediate transformations and often achieve improved performance and efficiency.

\subsection{BEV Semantic Segmentation}

BEV semantic segmentation aims to predict dense semantic maps in bird’s-eye view, assigning a semantic label to each spatial grid cell.
Early works~\citep{LSS,fiery,yang2021projecting,roddick2020predicting,VPN} adopt pipelines similar to BEVDet~\citep{BEVDet}, differing mainly in task-specific prediction heads.
VPN~\citep{VPN} leverages synthetic 3D environments with domain adaptation for real-world deployment.
M2BEV~\citep{xie2022m} transforms multi-view image features into unified BEV representations within ego-centric coordinates, enabling multi-task learning, while CVT~\citep{zhou2022cross} employs cross-view attention with camera-aware positional embeddings to learn perspective-to-BEV mappings.
HDMapNet~\citep{li2022hdmapnet} further extends BEV representations by predicting vectorized map elements directly in BEV space.

Unlike 3D object detection, which primarily targets foreground instances, BEV semantic segmentation focuses on dense background map elements.
The task typically decodes semantic categories directly from BEV feature maps, making BEV representations the central 3D abstraction in most BEV semantic segmentation frameworks.

\begin{figure*}[ht]
\setlength{\abovecaptionskip}{10pt}
    \centering
    \includegraphics[width=\linewidth]{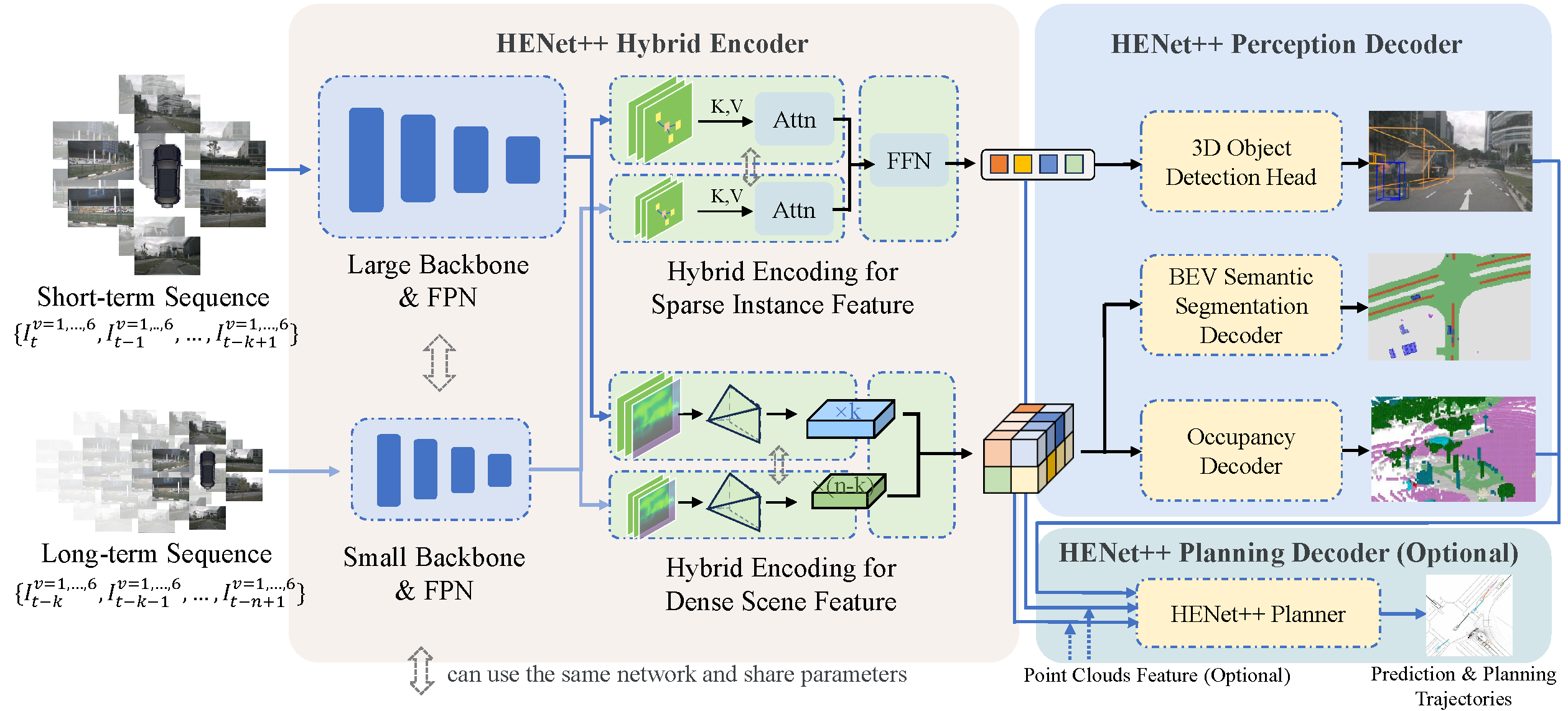}

\caption{\textbf{Overall architecture of \ourspp.}
The proposed framework performs hybrid encoding to jointly learn sparse foreground representations and dense background voxel features, enabling end-to-end multi-task prediction. A model-merging-based pre-training strategy is further introduced to improve multi-task performance.
}
    \label{fig:arch222}
\end{figure*}

\subsection{Occupancy Prediction}

Occupancy perception represents the environment as discretized 3D grids and predicts whether each voxel is occupied, with later extensions incorporating semantic labels for semantic occupancy prediction.
Early work such as MonoScene~\citep{22} demonstrates occupancy prediction using monocular RGB inputs, while TPVFormer~\citep{23} leverages multi-view cameras with transformer-based tri-perspective representations.
Subsequent approaches explore different strategies for efficient 3D representation learning, including spatial cross-attention from BEV features (SurroundOcc~\citep{24}), two-stage transformer frameworks for semantic scene completion (VoxFormer~\citep{25}), and efficient BEV-to-3D lifting mechanisms (FlashOcc~\citep{26}).
Other works improve view transformation and representation efficiency, such as front-to-back projection (FBOcc~\citep{27}) and compact occupancy modeling with coarse-to-fine semantic grouping (COTR~\citep{32}).
NeRF-based approaches (UniOcc~\citep{28}, RenderOcc~\citep{29}) model occupancy via implicit representations but are constrained by rendering efficiency, while FastOcc~\citep{31} focuses on accelerating inference through improved prediction heads.

Occupancy prediction models both foreground objects and background structures, capturing richer scene information than BEV semantic segmentation.
By nature, it can be viewed as a dense 3D semantic segmentation task.
Consequently, unlike object detection, which benefits from sparse representations, occupancy prediction typically favors dense representations such as BEV or voxel-based features.

\subsection{End-to-end Multi-task Learning}

End-to-end multi-task learning aims to jointly solve multiple perception tasks within a unified framework, reducing redundant computation and improving system efficiency.
Existing work explores joint learning of 3D object detection and BEV semantic segmentation~\citep{PMF,OCNet,roddick2020predicting,VPN}, often following BEVFormer~\citep{BEVFormer} by generating a shared BEV representation for multiple tasks.
Alternative designs, such as PETRv2~\citep{PETRv2}, employ separate query sets for different tasks, revealing inherent conflicts between sparse object-centric detection and dense BEV-based segmentation.

Beyond perception, end-to-end autonomous driving integrates perception, prediction, and planning into a unified pipeline~\citep{hu2022st}.
Representative approaches include BEV-based feature sharing with complex task interactions (UniAD~\citep{UniAD}), vectorized scene representations (VAD~\citep{jiang2023vad}), and simplified task coupling to reduce error accumulation (PARA-Drive~\citep{weng2024drive}).
Recent works explore different representations and planning strategies, including hybrid dense-sparse pipelines (GenAD~\citep{zheng2024genad}), purely sparse representations (SparseDrive~\citep{sun2024sparsedrive}), and advanced planning modules such as momentum-aware planning, trajectory bridging, and diffusion-based generation~\citep{song2025don,zhang2025bridging,liao2025diffusiondrive}.

Purely sparse representations focus on foreground elements and may lack comprehensive scene understanding, while purely dense representations require additional processing to extract object instances, potentially introducing errors.
Recent methods actively mitigate negative transfer during multi-task learning.
MMTL-UniAD~\citep{liu2025mmtl} dynamically decides which parameters to share and which to keep task-specific.
Graph Coloring for Multi-Task Learning~\citep{patapati2025graph} avoids the simultaneous updating of conflicting tasks.
RotoGrad~\citep{javaloy2021rotograd} employs gradient homogenization strategies, and ForkMerge~\citep{jiang2023forkmerge} employs dynamic branching and merging techniques that allow the network to temporarily diverge during training to explore task-specific solutions before reconciling them.
Our method integrates dense scene representations, guided by occupancy prediction, with sparse object-centric features from 3D detection, enabling more comprehensive perception for downstream prediction and planning.

\begin{figure*}[tp]
   \setlength{\abovecaptionskip}{10pt}
    \centering
    \includegraphics[width=\linewidth]{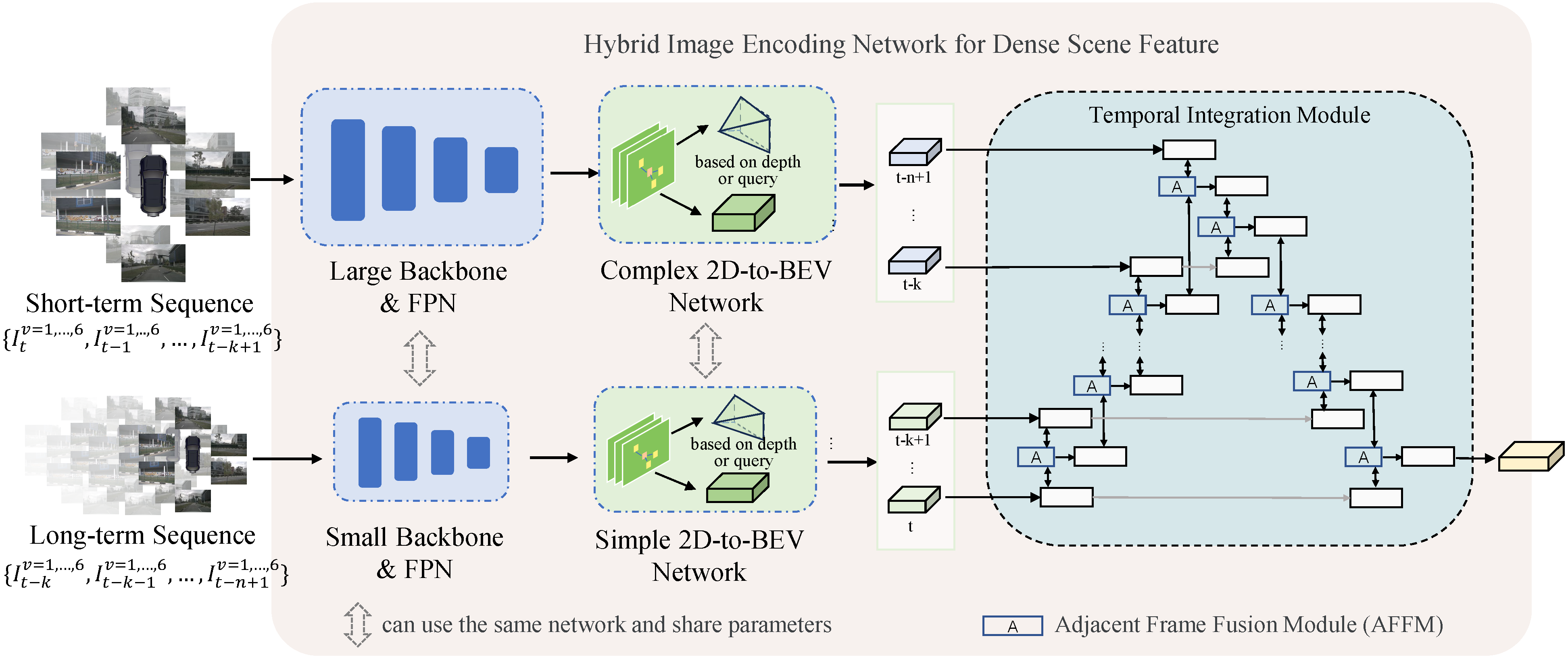}
    \caption{\textbf{Design of Hybrid Image Encoding Network for dense feature extraction.} 
    Hybrid Image Encoding Network uses image encoders of varying complexity to encode long-sequence frames and short-term images, respectively. Temporal Integration Module fuses multi-frame features from the various encoders.}
    \label{fig:arch}
\end{figure*}

\section{Method}


\subsection{Overall framework of HENet++}


Analysis of task characteristics in prior work reveals potential sources of conflict in multi-task learning.
Tasks such as 3D object detection produce inherently sparse outputs, where object-centric representations must ultimately be extracted even when dense features are employed.
Intermediate steps, including depth estimation, feature projection, and heatmap clustering, may introduce cumulative errors, making sparse query-based architectures more suitable for such tasks.
In contrast, dense prediction tasks, such as BEV semantic segmentation and occupancy prediction, generate dense spatial outputs, and thus align more naturally with BEV or voxel-based representations.

To address these challenges and enhance multi-task accuracy with complementary foreground and background information, we propose HENet++, an extension of the HENet framework.
As illustrated in~\figref{fig:arch222}, HENet++ introduces a unified hybrid encoding scheme that simultaneously extracts sparse foreground features and dense panoramic representations.
Specifically, we extend hybrid encoding to support sparse and dense feature learning within a single framework.
Furthermore, a multi-task model consolidation-based pre-training strategy is adopted to further improve overall performance.


\subsection{Hybrid Image Encoding Network}
\label{sec:henet}

\subsubsection{Encoding Dense Scene Features}
\label{sec:henet1}

As illustrated in~\figref{fig:arch}, the hybrid image encoding network employs two image encoders with different complexities to process temporal inputs at different scales.
The first encoder focuses on high-resolution short-term frames, utilizing a large backbone (\textit{e.g.}, VoVNetV2-99~\citep{v299}) with a feature pyramid network (FPN)~\citep{FPN}, followed by a complex 2D-to-BEV transformation module to generate high-precision BEV features.
In our implementation, BEVStereo~\citep{BEVStereo} is adopted for this purpose.
The second encoder targets long-term temporal context by down-sampling inputs to lower resolution and applying a lightweight backbone (\textit{e.g.}, ResNet-50~\citep{resnet}) with FPN, followed by a simplified 2D-to-BEV module, implemented using BEVDepth~\citep{BEVDepth}.
Both branches employ BEVPoolv2~\citep{bevpoolv2} to project frustum features into multi-scale BEV representations.
Based on empirical analysis, BEV feature resolutions of $256\times256$ and $128\times128$ are used for 3D object detection and BEV semantic segmentation, respectively.
Certain components of the hybrid encoders can be shared; for example, a single backbone (\textit{e.g.}, ResNet-50) may be used with different 2D-to-BEV modules.


Following the extraction of multi-frame BEV features by the hybrid image encoding network, temporal fusion is performed using a temporal integration module, as illustrated in~\figref{fig:arch}.
The module consists of complementary backward and forward processes: the backward process propagates features from the current frame to past frames, while the forward process aggregates historical features into the current frame.
Each step employs a weight-sharing Adjacent Frame Fusion Module (AFFM), which uses cross-attention to fuse BEV features from adjacent frames.
These two processes correspond to the two loops in Algorithm~\ref{algorithm1}: the first loop propagates $f_0$ backward through the sequence, whereas the second loop aggregates the updated features forward. The algorithm finally returns the temporally integrated current-frame feature $f_0$ for subsequent dense prediction.

Given BEV features from two frames, $f_i$ and $f_j$, the AFFM operation is defined as:
\begin{equation}
\begin{split}
\text{AFFM}(f_i, f_j) = &\, f_j + \gamma \times \text{Avg}(
     Atn(\langle f_i, f_j \rangle, f_i, f_i), \\ &
     Atn(\langle f_i, f_j \rangle, f_j, f_j)),
\end{split}
\end{equation}
where $Avg(\cdot)$ denotes the averaging operator, $\gamma$ is a learnable scaling parameter, $\langle \cdot, \cdot \rangle$ represents concatenation, and the cross-attention operator is defined as:
\begin{equation}
Atn(q,k,v) = \mathrm{softmax}\left(\frac{qk^\top}{\sqrt{d}}\right)v.
\end{equation}
In the backward process, $j=i-1$, whereas in the forward process, $j=i+1$.
Adjacent attention reduces noise compared to global attention or cross-frame convolutions, enabling effective alignment of moving objects while suppressing redundant background information.

\begin{algorithm}[t]
\caption{\small{Pseudo-code for Temporal Integration Module }}
\label{algorithm1}
\begin{algorithmic}
\STATE \textbf{Input:} A series of BEV features $\{f_{-n+1}, ... , f_{-1}, f_0\}$. 
$f_0$ represents BEV feature of the current frame and $f_{-i}$ corresponds to the $i^{th}$ frame before $f_0$.
    
\FOR{i \textbf{from} 0 \textbf{to} n-2}
\STATE $f_{-(i+1)} \leftarrow \text{AFFM}(f_{-i}, f_{-(i+1)})$
\ENDFOR

\FOR{i \textbf{from} n-2 \textbf{to} 0}
\STATE $f_{-i} \leftarrow \text{AFFM}(f_{-(i+1)}, f_{-i})$
\ENDFOR

\textbf{return} $f_{0}$

\end{algorithmic}
\end{algorithm}

\subsubsection{Encoding Sparse Instance Features}

\begin{figure*}[tp]
    \setlength{\abovecaptionskip}{10pt}
    \centering
    \includegraphics[width=\linewidth]{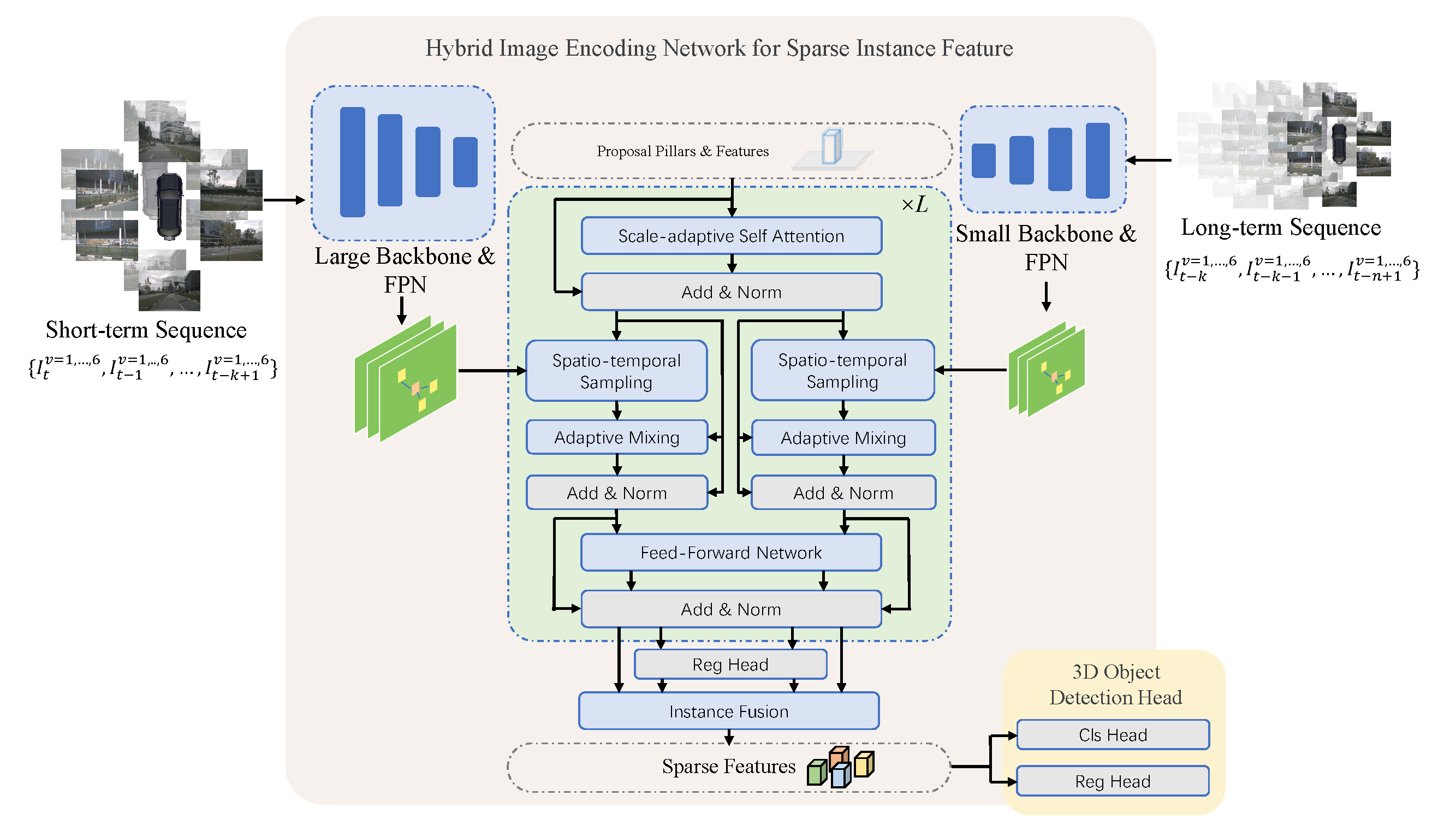}
\caption{\textbf{Hybrid image encoding network for sparse feature extraction.} The architecture extracts object-centric representations through hybrid encoding of multi-view image features.}
    \label{fig:hybridsp}
\end{figure*}


For sparse feature extraction, we build upon SparseBEV~\citep{SparseBEV} with several modifications.
We adopt learnable initialization and scale-adaptive self-attention to initialize queries $q \in \mathbb{R}^{N \times d}$:
\begin{equation}
q = \text{LayerNorm}\left( Q + \text{Softmax}\left(\frac{QK^\top}{\sqrt{d}} - \tau D\right) V \right),
\end{equation}
where $Q, K, V \in \mathbb{R}^{N \times d}$ denote the query representations, $N$ is the number of queries, $d$ is the channel dimension, and $\tau$ is a learnable scalar controlling the receptive field of each query.
The initialized queries are used to jointly attend to both short-term and long-term feature representations.

As shown in Figure~\ref{fig:hybridsp}, suppose that there are $k$ frames for short-term image features $f_{short}$ and $n-k$ frames for long-term features $f_{long}$, with $S$ sampling points per frame.
Two separate Spatio-temporal Sampling modules~\citep{SparseBEV} are employed to extract sampling features
$P_{short}\in\mathbb{R}^{k\times S\times C}$ and $P_{long}\in\mathbb{R}^{(n-k)\times S\times C}$, respectively.
These sampled features are then aggregated via Adaptive Mixing to obtain the short-term instance feature $q_{short}^{'}\in \mathbb{R}^{N \times d}$ and the long-term instance feature $q_{long}^{'} \in \mathbb{R}^{N \times d}$:
\begin{align}
\label{eq:1}
   \langle q_{short}, q_{long} \rangle & = \text{ReLU}(\text{LayerNorm}( \nonumber  
  \langle P_{short}, P_{long}\rangle \\ & \cdot\text{Linear}^{C\times C}(\langle q,q\rangle))), \\
\label{eq:2}
   q_{short}^{'} & = \text{ReLU}(\text{LayerNorm}( \nonumber   
  q_{short}^\top \\& \cdot\text{Linear}^{kS\times kS}(q))), \\
\label{eq:3}
   q_{long}^{'} & = \text{ReLU}(\text{LayerNorm}( \nonumber \\&  
  \qquad q_{long}^\top \cdot\text{Linear}^{(n-k)S\times (n-k)S}(q))),
\end{align}
The network weights for Channel Mixing (Equation~\ref{eq:1}) are shared between long-term and short-term features, while Point Mixing (Equations~\ref{eq:2} and~\ref{eq:3}) uses separate networks for the two feature types.
The resulting features, $q_{short}^{'}$ and $q_{long}^{'}$, are passed through a feed-forward network (FFN) before entering the next iteration of Self-Attention, Sampling, and Mixing.
This process is repeated for a total of $L$ iterations.

After $L$ iterations, $N$ short-term and $N$ long-term instance features are obtained.
Although these features originate from one-to-one corresponding queries $q$, the attended regions may shift during iterative attention, potentially leading to mismatched or overlapping instances.
Therefore, directly merging them into $N$ features or naively concatenating them into $2N$ features is suboptimal.

To solve this problem, we employ the regression head of the 3D object detection module to decode instance bounding boxes via an MLP. After that, a deduplication operation, analogous to non-maximum suppression, is performed to pinpoint features corresponding to the same instance.
For each instance, the associated $C$-dimensional feature vectors are fused into a single representation via channel-wise max pooling.


\begin{algorithm}[tp]
\caption{\small Model Merge for Pretraining based on~\citep{regmean}}
\label{regmean}
\begin{algorithmic}
\STATE \textbf{Input:} Single-task encoder $f_{1..K}$, Number of linear layers $J$, inner product matrices $G_i^{(j)}=X_i^{(j)T}X_i^{(j)}$ for all linear layers $1\leq j \leq J$ and single-task encoder $1\leq i \leq K$, Scaling factor of non-diagonal items $\alpha$

\FOR{$j$ $\mathbf{in}$ $1,2,...,J$}
\STATE $W_1^{(j)}, W_2^{(j)}..., W_K^{(j)} \leftarrow \textrm{getWeights}(f_{1..K}, j)$ \;
\STATE Reduce non-diagonal items of inner product matrices $G_i^{(j)}$ as $\tilde{G}_i^{(j)} \leftarrow \alpha G_i^{(j)} + (1-\alpha) \textrm{diag}(G_i^{(j)})$ \;
\STATE $W_M^{(j)} \leftarrow (\sum_i^{i \in \mathcal{K}} \tilde{G}_i^{(j)})^{-1} \sum_i^{i \in \mathcal{K}}(\tilde{G}_i^{(j)} W_i^{(j)})$ and set the weight as $W_M^{(j)}$ in $f_M$
\ENDFOR

Average weights as $W_M=\frac{1}{K}\sum_i^{i \in \mathcal{K}} W_i$ for weights other than linear weights in $f_M$

\end{algorithmic}
\end{algorithm}

\subsection{Perception Decoders and Losses}
\label{sec:henetpploss}

For 3D object detection, two separate MLP heads are used to predict object categories and bounding boxes from sparse instance features.
The classification loss is Focal Loss, denoted as $\mathcal{L}_{cls}$, and the bounding box regression loss is the $\ell_1$ loss, denoted as $\mathcal{L}_{bbox}$.

For BEV semantic segmentation, we employ a SegNet-based decoder~\citep{segnet} optimized with focal loss $\mathcal{L}_{seg}$.
For occupancy prediction, a lightweight MLP decoder is adopted with focal loss $\mathcal{L}_{occ}$.
Additionally, a binary cross-entropy loss $\mathcal{L}_{depth}$ is applied for depth estimation.

The overall training objective is defined as a weighted sum:
\begin{equation}
\begin{split}
\mathcal{L} = & \alpha_{depth} \mathcal{L}_{depth}
+ \alpha_{cls} \mathcal{L}_{cls}
+ \alpha_{bbox} \mathcal{L}_{bbox} \\
& + \alpha_{seg} \mathcal{L}_{seg}
+ \alpha_{occ} \mathcal{L}_{occ},
\end{split}
\end{equation}
where $\alpha_{depth}$, $\alpha_{cls}$, $\alpha_{bbox}$, $\alpha_{seg}$, and $\alpha_{occ}$ are balancing weights.


\subsection{Model Initialization}

Parameter initialization plays a critical role in training multi-task models.
In our experiments, initializing the encoder with parameters from a model pre-trained on a single 3D task, along with the corresponding decoder weights, leads to faster convergence and higher accuracy than using backbone and neck parameters pre-trained on 2D tasks.
However, since the multi-task framework employs a shared encoder, selecting an appropriate single-task model for pre-training becomes an important design consideration.
Empirical results show that initializing the encoder with a 3D object detection model improves detection performance, but degrades the accuracy of other tasks.
In contrast, initializing the model with an occupancy model improves occupancy prediction but degrades performance on the remaining tasks.

\begin{figure}[tp]
    \setlength{\abovecaptionskip}{10pt}
    \centering
    \includegraphics[width=\linewidth]{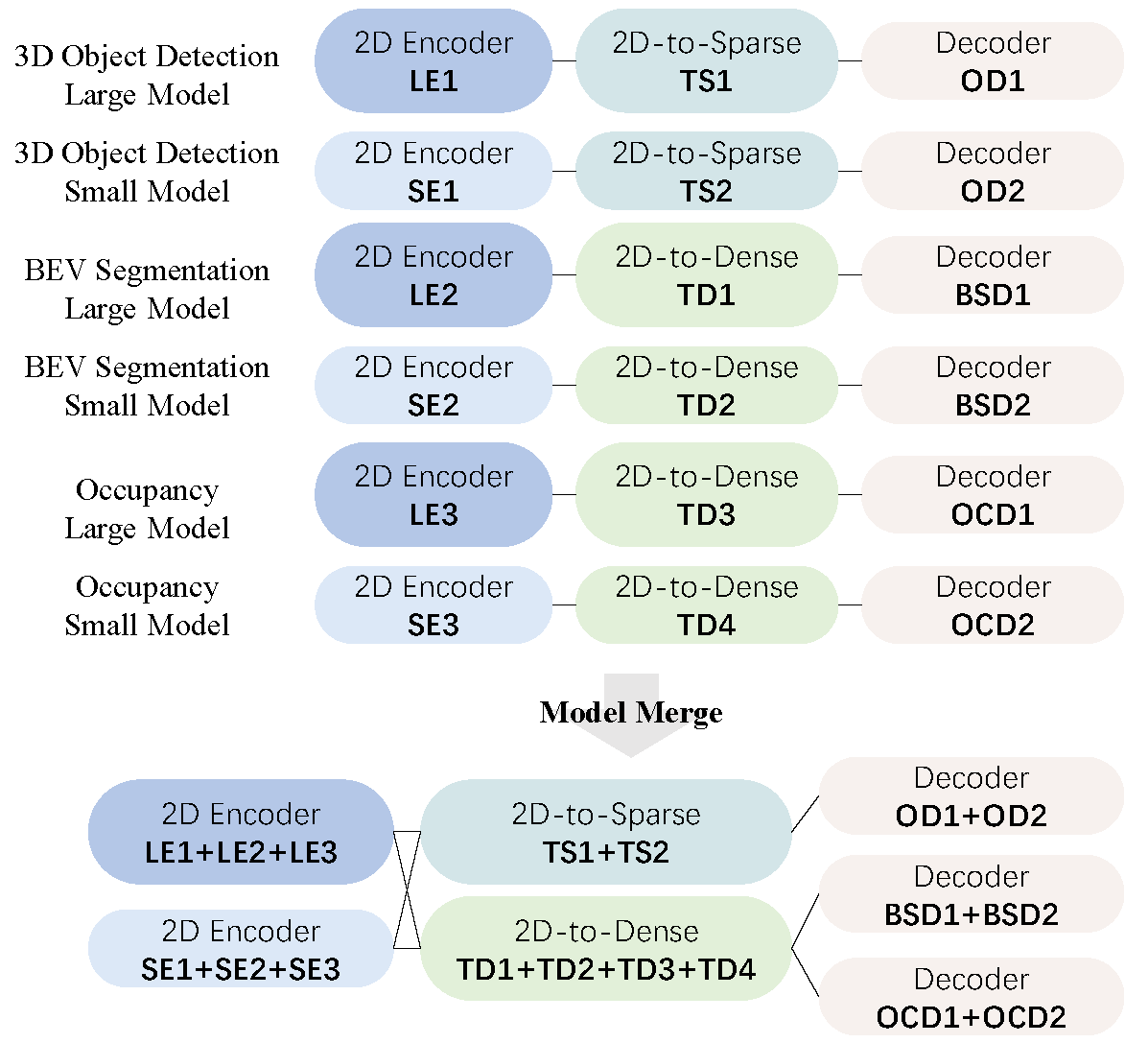}
    \caption{\textbf{Initialization of HENet++ via model merging.}}
    \label{fig:merge}
\end{figure}

Model merging, widely explored in natural language processing, aims to combine multiple single-task models into a unified model for multi-task inference.
We attempted to directly apply model merging methods, such as Regmean~\citep{regmean}, to 3D vision multi-task learning but observed limited performance gains, likely due to the complex architectures and heterogeneous task representations in 3D perception models.
Inspired by this idea, we instead consolidate the encoder weights from multiple single-task models into a unified encoder and use it as the pre-training initialization. This strategy provides an initialization that incorporates knowledge from 3D object detection, BEV semantic segmentation, and occupancy prediction, which yields promising performance improvements.

Algorithm~\ref{regmean} details this model-merging procedure. Given $K$ single-task encoders and their layer-wise inner-product matrices, Regression Mean is applied to the weight matrices of convolutional and linear layers, while the remaining learnable parameters are averaged. The resulting merged encoder $f_M$ is then used to initialize the unified encoder of HENet++ for subsequent multi-task training.
The HENet++ architecture supports three tasks, each comprising short-term and long-term branches inherited from models of different scales.
\figref{fig:merge} shows the multi-task model initialized by model merging. 




\begin{figure*}[tp]
    \setlength{\abovecaptionskip}{10pt}
    \centering
    \includegraphics[width=\linewidth]{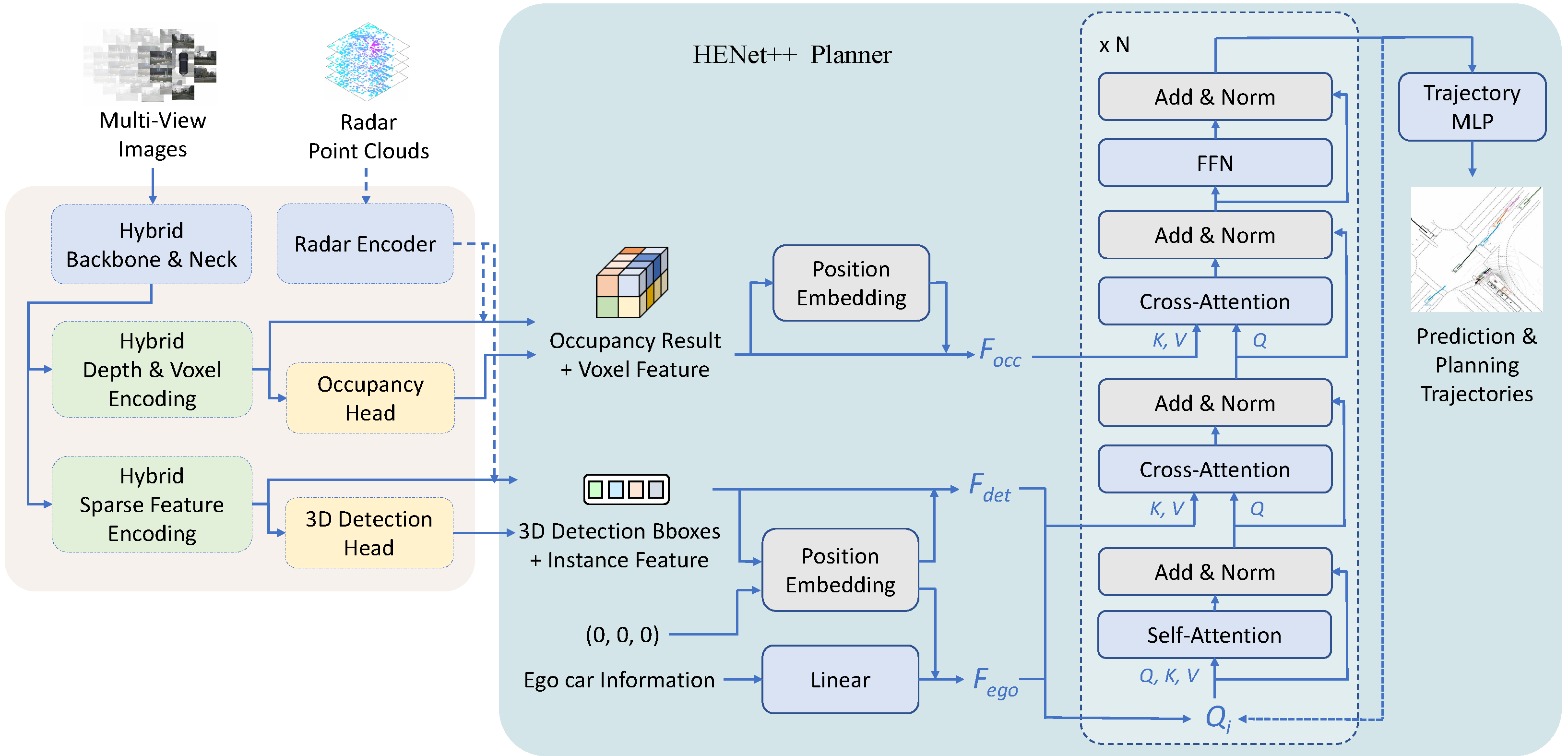}
\caption{\textbf{End-to-end autonomous driving architecture based on HENet++.}
An attention-based trajectory planner uses instance features (including the ego vehicle) as queries and combines instance and panoramic dense features as key-value pairs to enable iterative prediction of future states and planning.
The hybrid dense-sparse design of HENet++ further allows straightforward extension to multi-modal inputs.
}

    \label{fig:e2e}
\end{figure*}

\subsection{End-to-end Planning via HENet++}
\label{sec:e2e}

As illustrated in~\figref{fig:e2e}, we extend the HENet++ framework to end-to-end autonomous driving by introducing a simple yet effective iterative prediction and planning decoder.
Unlike existing approaches that rely on a single type of 3D representation, HENet++ leverages complementary sparse and dense features for planning.
Sparse instance features serve as learnable queries for object motion prediction, forming the basis for both prediction and planning, while dense panoramic features provide comprehensive scene context.

Specifically, the prediction and planning decoder in HENet++ consists of $N$ Transformer layers that perform iterative prediction.
Before iteration begins, the top-$K$ 3D object detection results with the highest confidence scores are selected together with their instance features $F_{det}$.
The positional encoding of their bounding box centroids is concatenated to form $K$ object vectors $F_{1,2,\ldots,K}$.
Ego-vehicle states, including historical trajectories, navigation waypoints, acceleration, and velocity, are projected into a higher-dimensional space via a linear layer and concatenated with the positional encoding of the origin to obtain the ego representation $F_{0}$.
Occupancy grid features are downsampled, and the semantic categories and voxel features are concatenated along the channel dimension with positional encoding of grid coordinates to produce occupancy tokens $F_{occ}$.

The initial query is defined as $Q_0 = \langle F_{0}, F_{1,2,\ldots,K} \rangle$.
At the $i$-th iteration, $Q_{i-1}$ serves as the input query, followed by self-attention on $Q_{i-1}$, cross-attention with $Q_0$ as key-value, cross-attention with $F_{occ}$ as key-value, and a feed-forward network to produce $Q_i$.
Each layer output is decoded by a trajectory MLP to predict trajectories $Trj_{0,1,\ldots,K}$, where $Trj_{0}$ denotes the planned ego trajectory and $Trj_{1,\ldots,K}$ correspond to predicted trajectories of surrounding objects.

The HENet++ framework is compatible with both sparse and dense representations, enabling straightforward integration with existing multi-modal fusion methods.
For example, by adopting the Radar encoder from RCBEVDet++~\citep{lin2024rcbevdet++} together with its interpolation strategy, both sparse and dense Radar features can be obtained.
These features can then be incorporated into the end-to-end model using the CAMF module proposed in~\citep{lin2024rcbevdet++}.

Denoting the perception loss defined in~\secref{sec:henetpploss} as $\mathcal{L}_{prec}$, the end-to-end model introduces three additional loss terms.
The first is an $\ell_1$ loss between the predicted ego trajectory and the ground truth, denoted as $\mathcal{L}_{plan}$.
The second is an $\ell_1$ loss for future trajectory prediction of surrounding objects, denoted as $\mathcal{L}_{pred}$.
Each trajectory $Trj_i$ consists of $t$ future points in the horizontal Cartesian coordinate system:
$\{(x_{i1}, y_{i1}), (x_{i2}, y_{i2}), \ldots, (x_{it}, y_{it})\}$.

The third term is a collision constraint $\mathcal{L}_{col}$, defined as:
\begin{equation}
\begin{split}
D(Trj_a, Trj_b, t_i) =
\sqrt{(x_{at_i}- x_{bt_i})^2 + (y_{at_i}- y_{bt_i})^2},
\end{split}
\end{equation}
\begin{equation}
\begin{split}
P(Trj_a, Trj_b, t_i) = \\
\begin{cases}
D_0 - D(Trj_a, Trj_b, t_i), & \text{if } D(Trj_a, Trj_b, t_i) < D_0, \\
0, & \text{otherwise},
\end{cases}
\end{split}
\end{equation}
\begin{equation}
\mathcal{L}_{col} = \sum_{i=1}^{K} \sum_{j=1}^\top P(Trj_0, Trj_i, t_j).
\end{equation}

The overall loss is a weighted sum:
\begin{equation}
\mathcal{L} = \mathcal{L}_{prec}
+ \alpha_{plan}\mathcal{L}_{plan}
+ \alpha_{pred}\mathcal{L}_{pred}
+ \alpha_{col}\mathcal{L}_{col},
\end{equation}
where $\alpha_{plan}$, $\alpha_{pred}$, and $\alpha_{col}$ are balancing weights.


\section{Experiments}
\label{sec:exp}

\subsection{Implementation Details}

\begin{table*}[t]
\setlength{\belowcaptionskip}{10pt}
\centering
\tablestyle{1.0pt}{1.2}
\caption[results1]{\textbf{Comparison of end-to-end multi-task learning on the nuScences {\tt{val}} set}. 
\textbf{Time/E} represent the training time per epoch with FP32 on 8$\times$A800 GPUs.
$\textbf{mIoU}_{\textbf{v}}$, $\textbf{mIoU}_{\textbf{a}}$, and $\textbf{mIoU}_{\textbf{d}}$ represent the mIoU for vehicles, drivable area, and lane \& road divider, respectively. 

}

\label{table:exp1} 
\begin{tabular}{l|cc|c||cc|cccc|cc}
\toprule
\multirow{2}{*}{\textbf{Methods}} & \multirow{2}{*}{\textbf{Backbone}}  & \multirow{2}{*}{\textbf{Frames}} & \multirow{2}{*}{\textbf{Time/E}}& \multicolumn{2}{c|}{\textbf{Detection}} & \multicolumn{4}{c|}{\textbf{BEV Segmentation}} & \multicolumn{2}{c}{\textbf{Occupancy}} \\
 &  &  &  & \textbf{NDS$\uparrow$} & \textbf{mAP$\uparrow$} & 
$\textbf{mIoU}\uparrow$ & 
$\textbf{IoU}_{\textbf{v}}\uparrow$ & $\textbf{mIoU}_{\textbf{a}}\uparrow$ & $\textbf{mIoU}_{\textbf{d}}\uparrow$ & $\textbf{IoU}\uparrow$ & $\textbf{mIoU}\uparrow$\\

\hline
{VPN} & ResNet50 &  1 & - & 33.4 & 25.7 & 43.8 & 37.3 & 76.0 & 18.0 & - & - \\
\gr {LSS} & ResNet50 &   1  & - & 41.0 & 34.4 & 45.0 & 42.8 & 73.9 & 18.3 & - & 11.4 \\
{BEVFormer-S} & ResNet101 &  1 & - & 45.3 & 38.0 & 47.3 & 44.4 & 77.6 & 19.8 & - & - \\

\hline
\gr {BEVFormer} & ResNet101 & 5 & 213min & 52.0 & 41.2 & 49.4 & 46.7 & 77.5 & 23.9 & - & 30.5 \\
{OccNet} & ResNet101 & 5 & 230min & 52.0 & 41.2 & 24.6 & 12.9 & 47.2 & 13.8 & 41.1 & 27.0 \\
\gr {UniAD} & ResNet101  & 6 & 253min & 49.9 & 38.2 & - & - & 69.1 & 25.7 & 62.3 & - \\
{PARA-Drive} & ResNet101  & 6 & 240min & 48.0 & 37.0 & - & - & 71.0 & 33.0 & 63.6 & - \\
\he \textbf{\ours~} & R101 \& R50  & 2 + 3 & 60min & 56.4 & 47.1 & 53.9 & 44.5 & 77.0 & 40.1 & - & -\\
\hepp \textbf{\ourspp~} & R101 \& R50  & 2 + 3 & 82min & 60.5 & 53.0 & 55.4 & 45.8 & 79.3 & 41.1 & 68.2 & 46.4 \\

\hline
{PETRv2} & V2-99 & 2 & 75min & 49.5 & 40.1 & 57.6 & 49.4 & 79.1 & \textbf{44.3} & - & - \\
\he \textbf{\ours~} & V2-99 \& R50 & 2 & 58min & 58.0 & 48.7 & 56.9 & 47.6 & 80.2 & 42.8 & - & - \\
\he \textbf{\ours~} & V2-99 \& R50 & 2 + 7 & 71min & 59.9 & 49.9 & 58.0 & 49.5 & 81.3 & 43.4 & - & - \\
\hepp \textbf{\ourspp~} & V2-99 \& R50 & 5 + 11 & 205min & \textbf{64.1} & \textbf{56.9} & \textbf{58.4} & \textbf{49.7} & \textbf{81.8} & 43.7 & \textbf{71.3} & \textbf{49.0} \\
\bottomrule
\end{tabular}
\end{table*}

\begin{table*}[t]
\footnotesize
\tablestyle{1.5pt}{1.2}
\caption{\textbf{Comparison of End-to-end Autonomous Driving results on nuScenes \tt{val} set.} L2 refers to the L2 error between the predicted trajectory and the ground-truth trajectory of human driving.}
\label{tab:e2eresult}
\begin{tabular}{l|cc|cccc|cccc|c|cccc|cccc}
\toprule

\multirow{3}{*}{Method} & \multirow{3}{*}{Input} & \multirow{3}{*}{Backbone} & \multicolumn{9}{c|}{UniAD Metrics} & \multicolumn{8}{c}{VAD/STP3 Metrics} \\
& & & \multicolumn{4}{c|}{L2 (m) $\downarrow$} & \multicolumn{4}{c|}{Collision Rate (\%) $\downarrow$} & CCR $\downarrow$ & \multicolumn{4}{c|}{L2 (m) $\downarrow$} & \multicolumn{4}{c}{Collision Rate (\%) $\downarrow$} \\
\cmidrule(lr){4-7} \cmidrule(lr){8-11} \cmidrule(lr){12-15} \cmidrule(lr){16-19}
& & & 1s & 2s & 3s & Avg. & 1s & 2s & 3s & Avg. & Avg. & 1s & 2s & 3s & Avg. & 1s & 2s & 3s & Avg. \\
\midrule

ST-P3           & C & EfficientNet-b4 & 1.72 & 3.26 & 4.86 & 3.28 & 0.44 & 1.08 & 3.01 & 1.51 &8.37& 1.33 & 2.11 & 2.90 & 2.11 & 0.23 & 0.62 & 1.27 & 0.71 \\
\gr OccNet      & C & ResNet101-DCN & 1.29 & 2.13 & 2.99 & 2.14 & 0.21 & 0.59 & 1.37 & 0.72 &-& -&-&-&-&-&-&-&- \\
UniAD           & C & ResNet101 & 0.48 & 0.96 & 1.65 & 1.03 & 0.05 & 0.17 & 0.71 & 0.31 &1.59& 0.45 & 0.70 & 1.04 & 0.73 & 0.62 & 0.58 & 0.63 & 0.61 \\
\gr Ego-MLP & C & ResNet50 & 0.15 & 0.32 & 0.59 & 0.35 & 0.00 & 0.27 & 0.85 & 0.37 &2.93& -&-&-&-&-&-&-&- \\
BEV-Planner & C & ResNet50 & 0.16 & 0.32 & 0.57 & 0.35 & 0.00 & 0.29 & 0.73 & 0.34 &3.16& -&-&-&-&-&-&-&- \\
\gr VAD-Tiny      & C & ResNet50 & 0.60 & 1.23 & 2.06 & 1.30 & 0.31 & 0.53 & 1.33 & 0.72 &-& 0.46 & 0.76 & 1.12 & 0.78 & 0.21 & 0.35 & 0.58 & 0.38 \\
VAD-Base     & C & ResNet101 & 0.54 & 1.15 & 1.98 & 1.22 & 0.04 & 0.39 & 1.17 & 0.53 &2.47& 0.41 & 0.70 & 1.05 & 0.72 & 0.07 & 0.17 & 0.41 & 0.22 \\
\gr PARA-Drive & C & ResNet50 & -&-&-&-&-&-&-&- &-& 0.25 & 0.46 & 0.74 & 0.48 & 0.14 & 0.23 & 0.39 & 0.25 \\
GenAD        & C & ResNet50 & 0.36 & 0.83 & 1.56 & 0.91 & 0.06 & 0.23 & 1.00 & 0.43 &-& 0.28 & 0.49 & 0.78 & 0.52 & 0.08 & 0.14 & 0.34 & 0.19 \\
\gr SparseDrive   & C & ResNet50 & 0.44 & 0.92 & 1.69 & 1.01 & 0.07 & 0.19 & 0.71 & 0.32 &-& 0.29 & 0.58 & 0.96 & 0.61 & 0.01 & 0.05 & 0.18 & 0.08 \\
MomAD            & C & ResNet50 & 0.43 & 0.88 & 1.62 & 0.98 & 0.06 & 0.16 & 0.68 & 0.30 &-& 0.31 & 0.57 & 0.91 & 0.60 & 0.01 & 0.05 & 0.22 & 0.09 \\
\gr SSR               & C & ResNet50 & 0.24 & 0.65 & 1.36 & 0.75 & 0.00 & 0.10 & 0.36 & 0.15 &-& 0.18 & 0.36 & 0.63 & 0.39 & 0.01 & 0.04 & 0.12 & 0.06 \\
BridgeAD-S & C & ResNet50 & -&-&-&-&-&-&-&- &-& 0.29 & 0.57 & 0.92 & 0.59 & 0.01 & 0.05 & 0.22 & 0.09 \\
\gr BridgeAD-B & C & ResNet101 & -&-&-&-&-&-&-&- &-& 0.28 & 0.55 & 0.92 & 0.58 & 0.00 & 0.04 & 0.20 & 0.08 \\
DiffusionDrive & C & ResNet50 & -&-&-&-&-&-&-&- &-& 0.27 & 0.54 & 0.90 & 0.57 & 0.03 & 0.05 & 0.16 & 0.08 \\
\gr SpaRC-AD & RC & ResNet50 & -&-&-&-&-&-&-&- &-& 0.24 & 0.47 & 0.79 & 0.50 & 0.01 & 0.06 & 0.20 & 0.09 \\

\hepp \textbf{\ourspp~}  & C  & ResNet50 & 0.41 & 1.27 & 2.63 & 1.44 & 0.02 & 0.10 & 0.39 & 0.17 &1.63& 0.25 & 0.56 & 1.02 & 0.61 & 0.01 & 0.05 & 0.12 & 0.06 \\
\hepp \textbf{\ourspp~}  & RC & ResNet50 & 0.39 & 1.11 & 2.36 & 1.29 & 0.00 & 0.06 & 0.33 & \textbf{0.13} &1.57& 0.24 & 0.50 & 0.91 & 0.55 & 0.01 & 0.03 & 0.10 & \textbf{0.05} \\
\bottomrule
\end{tabular}
\end{table*}

\begin{table}[t]
\centering
\caption{\textbf{Open-loop results with closed-loop metrics on NAVSIM.}}
\label{tab:navsim}
\tablestyle{1.5pt}{1.2}
\begin{tabular}{l|c|ccccc|c}
\toprule
Method & Mod. & NC$\uparrow$ & DAC$\uparrow$ & TTC$\uparrow$ & Comf.$\uparrow$ & EP$\uparrow$ & \textbf{PDMS}$\uparrow$ \\
\midrule
UniAD & C &  97.8 & 91.9 & 92.9 & \textbf{100.0} & 78.8 & 83.4 \\
\gr PARA-Drive & C & 97.9 & 92.4 & 93.0 & 99.8 & 79.3 & 84.0 \\
iPad & C & 98.6 & 98.3 & \textbf{94.9} & \textbf{100.0} & 88.0 & 91.7 \\
\hepp \ourspp & C & \textbf{98.9} & \textbf{98.7} & 94.2 & \textbf{100.0} & \textbf{88.3} & \textbf{92.6}\\ 
\bottomrule
\end{tabular}
\end{table}

\begin{figure}[tp]
    \setlength{\abovecaptionskip}{10pt}
    \centering
    \includegraphics[width=0.8\linewidth]{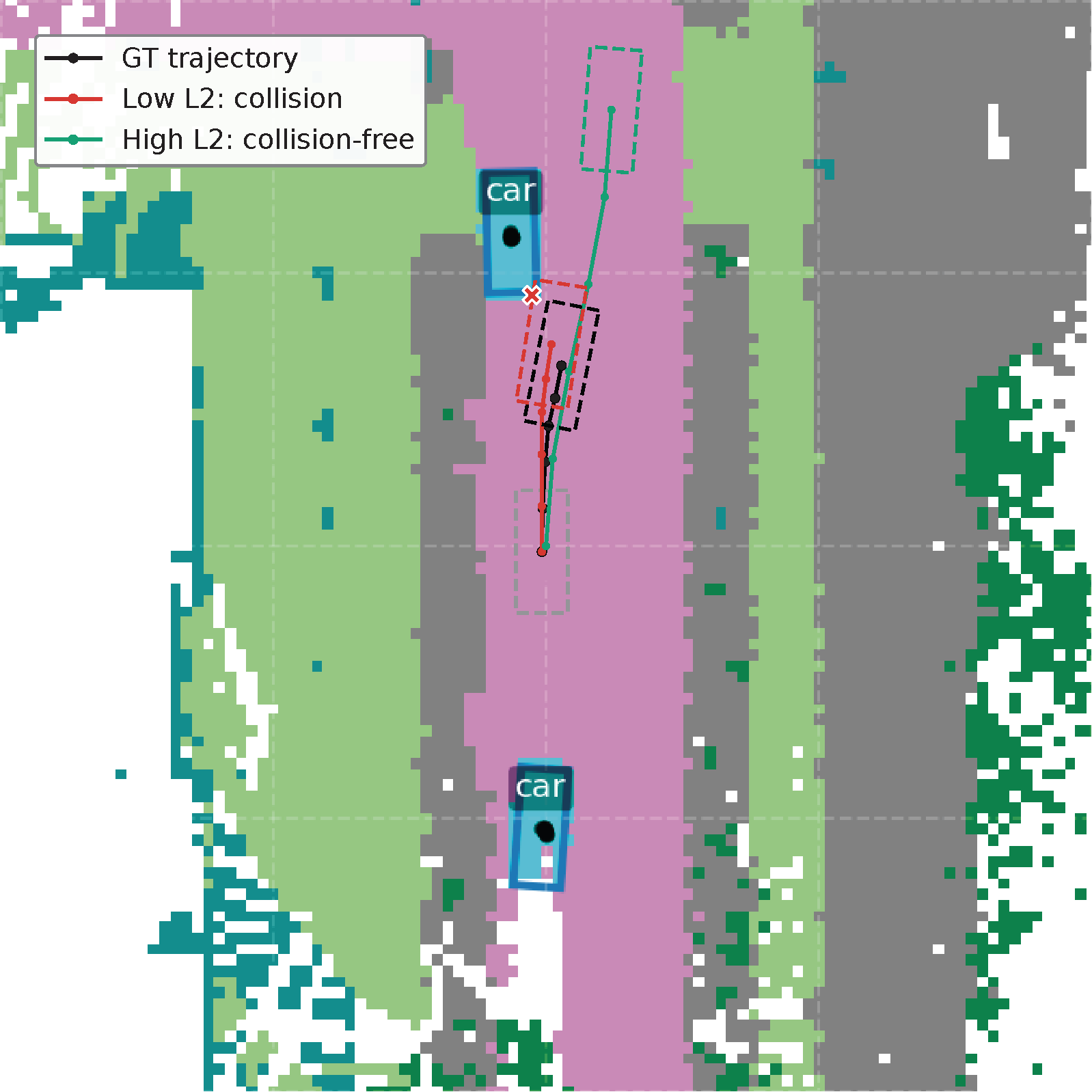}
    \caption{The baseline trajectory (red) has a lower L2 error with respect to the ground truth (black), but results in a collision. In contrast, HENet++ (green) produces a higher L2 error but successfully avoids collision.}
    \label{fig:colvsl2}
\end{figure}


\begin{table*}[t]
\setlength{\belowcaptionskip}{10pt}
\centering
\tablestyle{2pt}{1.2}
\caption[results2]{\textbf{Comparison of 3D object detection results on the nuScences {\tt{val}} set}. $*$ indicates the result benefits from the perspective pre-training. $\dagger$ indicates using one temporal frame. $\ddagger$ denotes integrating two or more temporal frames. The best and second best results are marked in \red{red} and \blue{blue}.}
\label{table:exp2} 
\begin{tabular}{l|c||cc|ccccc}
\toprule
{\textbf{Methods}} & \textbf{Backbone} & \textbf{NDS$\uparrow$} & \textbf{mAP$\uparrow$}  & \textbf{mATE$\downarrow$} & \textbf{mASE$\downarrow$} & \textbf{mAOE$\downarrow$} & \textbf{mAVE$\downarrow$} & \textbf{mAAE$\downarrow$}\\

\hline
{BEVDet} & ResNet50 & 37.9 & 29.8 & 0.725 & 0.279 & 0.589 & 0.860 & 0.245 \\
\gr {BEVDet4D$\dagger$}  & ResNet50 & 45.7 & 32.2 & 0.703 & 0.278 & 0.495 & 0.354 & 0.206 \\
{PETRv2$\dagger$}  & ResNet50 & 45.6 & 34.9 & 0.700 & 0.275 & 0.580 & 0.437 & 0.187 \\
\gr {BEVStereo$\dagger$} & ResNet50  & 50.0 & 37.2 & 0.598 & 0.270 & 0.438 & 0.367 & 0.190 \\
{SOLOFusion$\ddagger$} & ResNet50  & 53.4 & 42.7 & 0.567 & 0.274 & 0.511 & 0.252 & 0.181 \\
\gr {Sparse4Dv2$\ddagger$} & ResNet50  & 53.9 & \red{43.9} & 0.598 & 0.270 & 0.475 & 0.282 & 0.179 \\
{StreamPETR$\ddagger$} & ResNet50 & 54.0 & 43.2 & 0.581 & 0.272 & 0.413 & 0.295 & 0.195 \\
\gr {SparseBEV$\ddagger$} & ResNet50 & \blue{54.5} & 43.2 & 0.606 & 0.274 & 0.387 & 0.251 & 0.186 \\
\he \textbf{\ours~ $\ddagger$} & ResNet50  & \red{55.4} & \blue{43.7} & 0.512 & 0.262 & 0.367 & 0.285 & 0.213  \\

\hline
{PETR*} & ResNet101-DCN  & 44.1 & 36.6 & 0.717 & 0.267 & 0.412 & 0.834 & 0.190 \\
\gr {BEVDepth*$\dagger$} & ResNet101  & 53.5 & 41.2 & 0.565 & 0.266 & 0.358 & 0.331 & 0.190 \\
{BEVFormer*$\ddagger$} & ResNet101-DCN  & 51.7 & 41.6 & 0.673 & 0.274 & 0.372 & 0.394 & 0.198 \\
\gr {HoP-BEVFormer*$\ddagger$} & ResNet101-DCN & 55.8 & 45.4 & 0.565 & 0.265 & 0.327 & 0.337 & 0.194 \\
{StreamPETR*$\ddagger$} & V2-99  & 57.1 & 48.2 & 0.569 & 0.262 & 0.315 & 0.257 & 0.199 \\
\gr {SOLOFusion*$\ddagger$} & ResNet101 & 58.2 & 48.3 & 0.503 & 0.264 & 0.381 & 0.246 & 0.207 \\
{SparseBEV*$\ddagger$} & ResNet101  & 59.2 & 50.1 & 0.562 & 0.265 & 0.321 & 0.243 & 0.195\\
\gr {Far3D*$\ddagger$} & ResNet101  & 59.4 & \blue{51.0} & 0.551 & 0.258 & 0.372 & 0.238 & 0.195 \\
\he \textbf{\ours~ *$\ddagger$} & V2-99 \& ResNet50  & \blue{59.9} & 50.2 & 0.465 & 0.261 & 0.335 & 0.267 & 0.197 \\
\hepp \textbf{\ourspp~*$\ddagger$} & V2-99 \& ResNet50  & \red{65.1} & \red{57.3} & 0.506 & 0.252 & 0.182 & 0.225 & 0.195 \\
\bottomrule
\end{tabular}
\end{table*}

\begin{table*}[t]
\setlength{\belowcaptionskip}{10pt}
\centering
\tablestyle{1.5pt}{1.2}
\caption[results3]{\textbf{Comparison of 3D object detection results on nuScences {\tt{test}} set.} The best and second best results are marked in \red{red} and \blue{blue}. 
\ddag uses test-time augmentation.
}

\label{table:exp3} 
\begin{tabular}{l|c||cc|ccccc}
\toprule
{\textbf{Methods}} & \textbf{Backbone} & \textbf{NDS$\uparrow$} & \textbf{mAP$\uparrow$}  & \textbf{mATE$\downarrow$} & \textbf{mASE$\downarrow$} & \textbf{mAOE$\downarrow$} & \textbf{mAVE$\downarrow$} & \textbf{mAAE$\downarrow$}\\
\hline
{BEVDet4D}  & Swin-B  & 56.9 & 45.1 &0.511 & 0.241 & 0.386 & 0.301 & 0.121 \\
\gr {PolarFormer} & V2-99 & 57.2 & 49.3 & 0.556 & 0.256 & 0.364 & 0.439 & 0.127 \\
{PETRv2}  & V2-99  & 58.2 & 49.0 & 0.561 & 0.243 & 0.361 & 0.343 & 0.120 \\
\gr {HoP-BEVFormer} & V2-99  & 60.3 & 51.7 & 0.501 & 0.245 & 0.346 & 0.362 & 0.105 \\
{BEVDepth} & ConvNeXt-B  & 60.9 & 52.0 & 0.445 & 0.243 & 0.352 & 0.347 & 0.127 \\
\gr {BEVStereo} & V2-99  & 61.0 & 52.5 & 0.431 & 0.246 & 0.358 & 0.357 & 0.138 \\
{SOLOFusion} & ConvNeXt-B  & 61.9 & 54.0 & 0.453 & 0.257 & 0.376 & 0.276 & 0.148 \\
\gr {AeDet} & ConvNeXt-B  & 62.0 & 53.1 & 0.439 & 0.247 & 0.344 & 0.292 & 0.130 \\
{BEVFormerv2} & InternImage-B  & 62.0 & 54.0 & 0.488 & 0.251 & 0.335 & 0.302 & 0.122 \\
\gr {FB-BEV} & V2-99  & 62.4 & 53.7 & 0.439 & 0.250 & 0.358 & 0.270 & 0.128 \\
{StreamPETR} & V2-99  & \blue{63.6} & 55.0 & 0.479 & 0.239 & 0.317 & 0.241 & 0.119 \\
\gr {SparseBEV} & V2-99  & \blue{63.6} & \blue{55.6} & 0.485 & 0.244 & 0.332 & 0.246 & 0.117\\
{Sparse4Dv2} & V2-99  & \red{63.8} & \blue{55.6} & 0.462 & 0.238 & 0.328 & 0.264 & 0.115 \\
\he \textbf{\ours~} & V2-99 \& ResNet50  & \red{63.8} & \red{57.5} & 0.432 & 0.242 & 0.368 & 0.320 & 0.129 \\

\hline
{BEVFormerV2} & InternImage-XL & 64.8 & 58.0 & 0.448 & 0.262 & 0.342 & 0.238 & 0.128 \\
\gr {BEVDet-Gamma \ddag} & Swin-B & 66.4 & 58.6 & 0.375 & 0.243 & 0.377 & 0.174 & 0.123 \\
{SparseBEV} & V2-99 & 67.5 & 60.3 & 0.425 & 0.239 & 0.311 & 0.172 & 0.116 \\
\gr {StreamPETR-Large} & ViT-L & 67.6 & 62.0 & 0.470 & 0.241 & 0.258 & 0.236 & 0.134 \\ 
{Hop } & ViT-L & 68.5 & 62.4 & 0.367 & 0.249 & 0.353 & 0.171 & 0.131 \\
\gr {Far3D} & ViT-L & \blue{68.7} & \blue{63.5} & 0.432 & 0.237 & 0.278 & 0.227 & 0.130 \\
\hepp \textbf{\ourspp~} & ViT-L \& V2-99 & \red{70.7} & \red{64.5} & 0.402 & 0.235 & 0.237 & 0.155 & 0.129 \\
\bottomrule
\end{tabular}
\end{table*}


For HENet++, we adopt VoVNet-99~\citep{v299,dd3d} with $640\times1600$ image resolution as the large image encoder, and ResNet-50~\citep{resnet} with $256\times704$ resolution as the lightweight encoder, following~\citep{SparseBEV}.
For temporal inputs, the short-term and long-term frame numbers are set to $k=5$ and $n=11$, respectively.
In hybrid encoding for sparse features, the query number is $N=900$ and the number of layers is $L=6$.
For the hyperparameter $\alpha$ in model merging, we adopt $\alpha = 0.9$.

We additionally train three single-task models with hybrid encoding under the same settings as the multi-task model.
All models are trained on the nuScenes training set for 12 epochs without CBGS, except for the large detection model reported on the nuScenes test leaderboard, which uses an 8-frame ViT-L encoder and a 15-frame VoVNet-99.

For the HENet++ end-to-end autonomous driving model, the top 200 detection results are retained, and voxel features are downsampled to $50 \times 50 \times 4$.
The prediction and planning decoder uses $N=3$ Transformer layers.
The distance threshold $D_0$ in $\mathcal{L}_{col}$ is set to 3 meters.
After adding the prediction \& planning decoder to the perception model, the end-to-end system is fine-tuned for 10 epochs without CBGS.

\begin{table*}[t]
\setlength{\belowcaptionskip}{10pt}
\centering
\tablestyle{1.5pt}{1.2}
\caption[results3]{\textbf{Comparison of BEV semantic segmentation results on nuScences {\tt{val}} set.} The best and second best results are marked in \red{red} and \blue{blue}. For both the HENet and HENet++ frameworks, the single-task BEV segmentation models are identical.}

\label{table:exp4} 
\begin{tabular}{l|c|c|ccc}
\toprule
{\textbf{Methods}} & \textbf{Backbone} & $\textbf{mIoU}\uparrow$ & $\textbf{mIoU}_{\textbf{veh}}\uparrow$ & $\textbf{mIoU}_{\textbf{area}}\uparrow$ & $\textbf{mIoU}_{\textbf{div}}\uparrow$\\
\hline

\gr {VPN} & ResNet50 & 42.7 & 31.8 & 76.9 & 19.4 \\
{LSS} & ResNet50 & 46.5 & 41.7 & 77.7 & 20.0 \\
\gr {BEVFormer} & ResNet101-DCN & 50.2 & 44.8 & 80.1 & 25.7 \\
{FIERY} & ResNet-101 & - & 38.2 & - & - \\
\gr {M2BEV} & ResNeXt-101 & - & - & 77.2 & 40.5 \\
{PETRv2} & V2-99 & \red{60.3} & \blue{46.3} & \red{85.6} & \red{49.0} \\
\he \textbf{\ours/\ourspp~} & V2-99 \& ResNet50 & \blue{58.8} & \red{51.6} & \blue{82.3} & \blue{42.4} \\

\bottomrule
\end{tabular}
\end{table*}

\begin{table*}[t]
\footnotesize
\tablestyle{1.5pt}{1.2}
\caption{\textbf{Comparison of Occupancy results on the nuScenes-Occ3D benchmark.} 
We present the RayIoU and the mean IoU across categories, as well as the IoUs for different classes. The best and second best results are marked in \red{red} and \blue{blue}.
}
\label{tab:occresult}
\begin{tabular}{l|c|c|c|c| c c c c c c c c c c c c c c c c c}
\toprule
Method
& \rotatebox{90}{Backbone} & \rotatebox{90}{Visible Mask}  & RayIoU & mIoU$\uparrow$
& \rotatebox{90}{others}
& \rotatebox{90}{barrier}
& \rotatebox{90}{bicycle}
& \rotatebox{90}{bus}
& \rotatebox{90}{car}
& \rotatebox{90}{const. veh.}
& \rotatebox{90}{motorcycle}
& \rotatebox{90}{pedestrian}
& \rotatebox{90}{traffic cone}
& \rotatebox{90}{trailer}
& \rotatebox{90}{truck}
& \rotatebox{90}{drive. suf.}
& \rotatebox{90}{other flat}
& \rotatebox{90}{sidewalk}
& \rotatebox{90}{terrain}
& \rotatebox{90}{manmade}
& \rotatebox{90}{vegetation}
\\
\hline

BEVFormer & R-50 & \ding{52} & 32.4 & 23.7 & 5.0 & 38.8 & 10.0 & 34.4 & 41.1 & 13.2 & 16.5 & 18.2 & 17.8 & 18.7 & 27.7 & 49.0 & 27.7 & 29.1 & 25.4 & 15.4 & 14.5 \\
SparseOcc & R-50 & \ding{52} & 36.1 & 30.9 & - & - & - & - & - & - & - & - & - & - & - & - & - & - & - & - & - \\
OPUS & R-50 & \ding{52} & 38.4 & 33.2 & 10.7 & 39.8 & 21.3 & 39.8 & 45.3 & 23.4 & 21.8 & 17.8 & 19.3 & 27.5 & 33.2 & 71.6 & 37.1 & 45.1 & 43.6 & 33.8 & 33.2 \\
TPVFormer & R-50 & \ding{52} & - & 34.2 & 7.7 & 44.0 & 17.7 & 40.9 & 47.0 & 15.1 & 20.5 & 24.7 & 24.7 & 24.3 & 29.3 & 79.3 & 40.7 & 48.5 & 49.4 & 32.6 & 29.8 \\
ODG & R-50 & \ding{52} & \red{39.2} & 35.5 & \blue{13.7} & 39.0 & 23.0 & \blue{46.8} & 49.3 & 25.8 & 23.6 & 20.7 & 18.5 & 30.0 & 35.6 & 76.8 & 39.3 & 45.0 & 46.8 & 37.5 & 32.2 \\
OccFormer  & R-50 & \ding{52} & - & 37.4 & 9.2 & 45.8 & 18.2 & 42.8 & 50.3 & 24.0 & 20.8 & 22.9 & 21.0 & 31.9 & 38.1 & 80.1 & 38.2 & 50.8 & 54.3 & \blue{46.4} & \blue{40.2} \\
FBOcc &  R-50 & \ding{52} & 33.5 & \blue{42.1} & \red{14.3} & \blue{49.7} & \red{30.0} & 46.6 & \blue{51.5} & \blue{29.3} & \red{29.1} & \blue{29.4} & \red{30.5} & \blue{35.0} & \blue{39.4} & \blue{83.1} & \red{47.2} & \blue{55.6} & \red{59.9} & 44.9 & 39.6  \\ 

\hepp \textbf{\ourspp~} & R-50 & \ding{52} & \blue{38.9} & \red{42.9} & 10.8 & \red{50.3} & \blue{24.3} & \red{49.0} & \red{57.3} & \red{29.4} & \blue{24.4} & \red{30.1} & \blue{28.5} & \red{36.5} & \red{43.0} & \red{84.0} & \blue{43.1} & \red{56.0} & \blue{59.3} & \red{54.2} & \red{49.2} \\ 

\hline

SurroundOcc  & R-101 & \ding{52} & - & 37.1 & 9.0 & 46.3 & 17.1 & 46.5 & 52.0 & 20.1 & 21.5 & 23.5 & 18.7 & 31.5 & 37.6 & 81.9 & 41.6 & 50.8 & 53.9 & 42.9 & 37.2 \\
VoxFormer & R-101 & \ding{52} & 38.8 & 40.7 & - & - & - & - & - & - & - & - & - & - & - & - & - & - & - & - & - \\
FastOcc & R-101 & \ding{52} & - & 40.8 & 12.9 & 46.6 & 29.9 & 46.1 & 54.1 & 23.7 & 31.1 & 30.7 & 28.5 & 33.1 & 39.7 & 83.3 & 44.7 & 53.9 & 55.5 & 42.6 & 36.5 \\ 
PanoOcc  & R-101 & - & - &  42.1 & 11.7 & 50.5 & 29.6 & 49.4 & 55.5 & 23.3 & 33.3 & 30.6 & 31.0 & 34.4 & 42.6 & 83.3 & 44.2 & 54.4 & 56.0 & 45.9 & 40.4 \\ 
BEVDet4D  & Swin-B & \ding{52} & 32.6 & 42.5 & 12.4 & 50.2 & 27.0 & 51.9 & 54.7 & 28.4 & 29.0 & 29.0 & 28.3 & 37.1 & 42.5 & 82.6 & 43.2 & 54.9 & 58.3 & 48.8 & 43.8\\
FlashOcc & Swin-B & \ding{52} & - & 43.5 & 13.3 & 51.6 & 28.1 & 50.9 & 55.7 & 27.5 & 31.1 & 30.0 & 29.2 & 38.9 & 43.7 & 83.9 & 45.6 & 56.3 & 59.0 & 50.6 & 44.6\\
COTR & Swin-B & \ding{52} & - & 46.2 & \red{14.9} & 53.3 & 35.2 & 50.8 & 57.3 & \blue{35.4} & 34.1 & 33.5 & 37.1 & 39.0 & 45.0 & 84.5 & 48.7 & 57.6 & 61.1 & 51.6 & \blue{46.7}\\
FBOcc & V2-99 & \ding{52} & \blue{39.1} & \blue{48.9} & \blue{14.3} & \blue{57.0} & \red{38.3} & \blue{57.7} & \red{62.1} & 34.4 & \blue{39.4} & \blue{38.8} & \blue{39.4} & \blue{42.9} & \blue{50.0} & \blue{86.0} & \blue{50.2} & \red{60.1} & \blue{62.5} & \blue{52.4} & 45.7\\

\hepp \textbf{\ourspp~} & V2-99 \& R-50 & \ding{52} & \red{41.2} & \red{49.8} & 14.1 & \red{57.6} & \blue{38.2} & \red{60.1} & \blue{61.3} & \red{36.1} & \red{39.8} & \red{39.3} & \red{41.6} & \red{43.9} & \red{50.6} & \red{88.3} & \red{51.3} & \blue{59.4} & \red{63.4} & \red{53.8} & \red{48.1} \\  

\bottomrule
\end{tabular}
\end{table*}

\subsection{Dataset and Metrics}
We evaluate our perception model on the nuScenes~\citep{nuScenes} dataset.
nuScenes is a large-scale autonomous driving benchmark containing 1,000 driving scenes (700 for training, 150 for validation, and 150 for testing) covering diverse environments such as urban areas, highways, and rural roads.
Each scene includes multiple object categories, including vehicles, pedestrians, and bicycles.
For 3D object detection, nuScenes provides standard evaluation metrics, including mean Average Precision (mAP) and five true-positive (TP) metrics: ATE, ASE, AOE, AVE, and AAE, which measure translation, scale, orientation, velocity, and attribute errors, respectively.
Overall performance is summarized by the nuScenes Detection Score (NDS), a composite metric derived from these measures.
For BEV semantic segmentation, we follow LSS~\citep{LSS} and report mean Intersection-over-Union (mIoU).
For occupancy prediction, binary occupancy is evaluated using IoU, while semantic occupancy is assessed using class-averaged mIoU. We also report RayIoU~\citep{tang2024sparseocc} to better assess the depth and detail quality of occupancy predictions.

For end-to-end autonomous driving, we evaluate performance on the nuScenes dataset and NAVSIM.
The nuScenes dataset uses L2 distance to expert trajectories and collision rate.
Since prior works such as UniAD~\citep{UniAD}, VAD~\citep{jiang2023vad}, and ST-P3~\citep{hu2022st} adopt different evaluation protocols (e.g., temporal averaging and collision definitions), we report results under both benchmarks for fair comparison.
NAVSIM~\citep{dauner2024navsim} is a benchmark that evaluates autonomous driving planning via non‑reactive simulation, using the PDM Score as the primary metric. This composite score combines safety factors (collision avoidance, drivable area compliance) with weighted terms for comfort and progress toward the goal.

\subsection{Multi-task Results}

We compare HENet and HENet++ with prior end-to-end multi-task models on the nuScenes \texttt{val} set, as shown in~\tabref{table:exp1} and~\figref{fig:intro}.
HENet++ demonstrates strong multi-task performance and achieves state-of-the-art results across multiple benchmarks.
Specifically, HENet++ surpasses BEVFormer~\citep{BEVFormer} by 12.1 NDS and 15.7 mAP in 3D object detection, 9.0 mIoU in BEV semantic segmentation, and 18.5 mIoU in occupancy prediction.
Compared with PETRv2~\citep{PETRv2}, which achieves strong BEV segmentation performance, our method improves 3D object detection by 14.6 NDS while maintaining competitive segmentation accuracy.

During the experiments, we updated the decoder of the occupancy single‑task model from BEVDet4D to FB‑Occ, which improved the single‑task mIoU from 48.2 to 49.8. Correspondingly, the multi‑task results improved from 63.7 NDS for detection, 58.3 mIoU for segmentation, and 47.3 mIoU for occupancy to 64.1 NDS, 58.4 mIoU, and 49.0 mIoU, respectively. These results also demonstrate that using a more accurate single‑task model can accordingly improve the performance of the corresponding task in the multi‑task setting, and that the HENet++ training paradigm can slightly boost the performance of other tasks as well.

We further compare training efficiency by measuring per-epoch training time under the same batch size.
Hybrid Encoding achieves superior efficiency with reduced training time compared to existing methods.
Furthermore, it supports longer temporal inputs to improve performance, whereas approaches such as PETRv2 cannot process nine frames due to GPU memory constraints.

We evaluate HENet++ for end-to-end autonomous driving against prior methods in~\tabref{tab:e2eresult} and~\tabref{tab:navsim}.
HENet++ achieves a lower collision rate on the nuScenes planning benchmark.
Besides, HENet++ achieves higher PDMS than existing end-to-end autonomous driving models on the NAVSIM benchmark.
\figref{fig:colvsl2} shows a visualization result on nuScenes. Although the planned trajectory of HENet++ has a larger L2 distance from the ground truth, its decision is correct because it avoids collision. This also explains why in~\tabref{tab:e2eresult}, HENet++ achieves a lower collision rate despite having a higher L2 error.

\begin{table*}[t]
\setlength{\belowcaptionskip}{10pt}
\centering
\tablestyle{1.3pt}{1.2}
\caption{\textbf{Ablation of Hybrid Image Encoding Network on Detection Task.} 
`Time' denotes the overall training time.
`BS' indicates the batch size.
For A$\sim$F, FPS is measured on RTX3090 GPU with FP32. 
For H$\sim$I, FPS is measured on A100-SMX4 GPU with FP32 and FP16 mixed precision. 
For A$\sim$F, Training cost (GPU memory and Time) is estimated on 8$\times$Tesla A800 GPUs with FP32.
For H$\sim$I, Training cost is estimated on 8$\times$A100-SMX4 GPUs with FP32 and FP16 mixed precision.
}

\label{table:abl1} 
\begin{tabular}{c|l|ccc||cc|ccc}
\toprule
& \textbf{Model} & \textbf{Backbone} & \textbf{Input} & \textbf{Frames} & \textbf{NDS} & \textbf{mAP} & \textbf{FPS} & \textbf{GPU memory} & \textbf{Time}\\

\hline
\gr A & BEVDepth4D & R18 & 640$\times$1152 & 2 frames & 48.6 & 34.8 & 14.3 & 31.2G / BS=64 & 14h \\
B & BEVDepth4D & R18 & 256$\times$704 & 9 frames & 48.6 & 34.9 & 21.7 & 22.9G / BS=64 & 14h \\
\gr C & BEVDepth4D & R18 & 640$\times$1152 & 9 frames & 51.2 & 39.2 & 14.1 & 43.5G / BS=64 & 44h \\
A+B & Model Ensemble &-&-&- & 48.9 & 35.2 & 7.93 & - & - \\
\he A+B & Dense Hybrid Encoding & R18 \& R18 & 640$\times$1152 \& 256$\times$704 & 2 + 7 frames & 52.1 & 39.8 & 8.91 & 41.3G / BS=64 & 13h \\

\hline
\gr D & BEVDepth4D & R50 & 256$\times$704 & 9 frames & 53.8 & 40.9 & 19.1 & 14.1G / BS=16 & 16h \\
E & BEVStereo & V2-99 & 640$\times$1152 & 2 frames & 58.2 & 48.0 & 4.52 & 37.0G / BS=16 & 48h \\
\gr F & BEVStereo & V2-99 & 896$\times$1600 & 2 frames & 59.2 & 50.0 & 2.69 & 68.4G / BS=16 & 96h \\
D+E & Model Ensemble &-&-&- & 57.3 & 46.4 & 3.53 & - & - \\
\he D+E & Dense Hybrid Encoding w/o AFFM & V2-99 \& R50 & 640$\times$1152 \& 256$\times$704 & 2 + 7 frames & 59.2 & 49.9 & 3.71 & 42.3G / BS=16  & 27h \\
\he D+E & Dense Hybrid Encoding & V2-99 \& R50 & 640$\times$1152 \& 256$\times$704 & 2 + 7 frames & 59.9 & 50.2 & 3.65 & 43.5G / BS=16 & 27h\\

\hline
\gr H & SparseBEV & R50 & 256$\times$704 & 11 frames & 55.8 & 45.2 & 31.1 & 14.7G / BS=8 & 17h \\
I & SparseBEV & V2-99 & 640$\times$1600 & 8 frames & 63.2 & 55.1 & 9.6 & 70.4G / BS=8 & 100h \\
\gr H+I & Model Ensemble &-&-& 11 frames & 60.8 & 52.7 & 7.3 & - & - \\
\hepp H+I & Sparse Hybrid Encoding & V2-99 \& R50 & 640$\times$1600 \& 256$\times$704 & 5 + 6 frames & 65.1 & 57.3 & 12.1 & 68.9G / BS=8 & 72h\\

\bottomrule
\end{tabular}
\end{table*}





\begin{table*}[t]
\setlength{\belowcaptionskip}{10pt}
\centering
\tablestyle{1.5pt}{1.2}
\caption{\textbf{Ablation study of the Hybrid Image Encoding Network in the multi-task setting.}
$\textbf{mIoU}_{bev}$ and $\textbf{mIoU}_{occ}$ denote the mIoU for BEV semantic segmentation and semantic occupancy, respectively.}

\label{table:abl1b} 
\begin{tabular}{c|l|ccc|ccc}

\toprule
& \textbf{Model} & \textbf{Backbone} & \textbf{Input} & \textbf{Frames} & \textbf{NDS} & $\textbf{mIoU}_{bev}$ & $\textbf{mIoU}_{occ}$ \\

\hline
\gr A & BEVDepth4D & R50 & 256$\times$704 & 9 frames & 53.0 & 51.5 &- \\
B & BEVStereo & V2-99 & 640$\times$1152 & 2 frames & 58.0 & 55.8 &-\\
\gr C & BEVStereo & V2-99 & 896$\times$1600 & 2 frames & 58.9 & 56.7&-\\
A+B & Dense Hybrid Encoding w/o AFFM & V2-99 \& R50 & 640$\times$1152 \& 256$\times$704 & 2 + 7 frames & 59.0 & 56.9&-\\
\he A+B & Dense Hybrid Encoding & V2-99 \& R50 & 640$\times$1152 \& 256$\times$704 & 2 + 7 frames & 59.9 & 58.0&-\\

\hline
\gr D & \ourspp~w/o Hybrid Encoding & R50 & 256$\times$704 & 9 frames & 54.1 & 51.9 & 39.4 \\
E & \ourspp~w/o Hybrid Encoding & V2-99 & 640$\times$1600 & 8 frames & 63.0 & 58.1 & 45.5 \\
\hepp D+E & \ourspp~ & V2-99 \& R50 & 640$\times$1600 \& 256$\times$704 & 5 + 6 frames & 63.7 & 58.3 & 47.3 \\

\bottomrule
\end{tabular}
\end{table*}

\begin{figure*}[t]
    \setlength{\abovecaptionskip}{10pt}
    \centering
    \includegraphics[width=\linewidth]{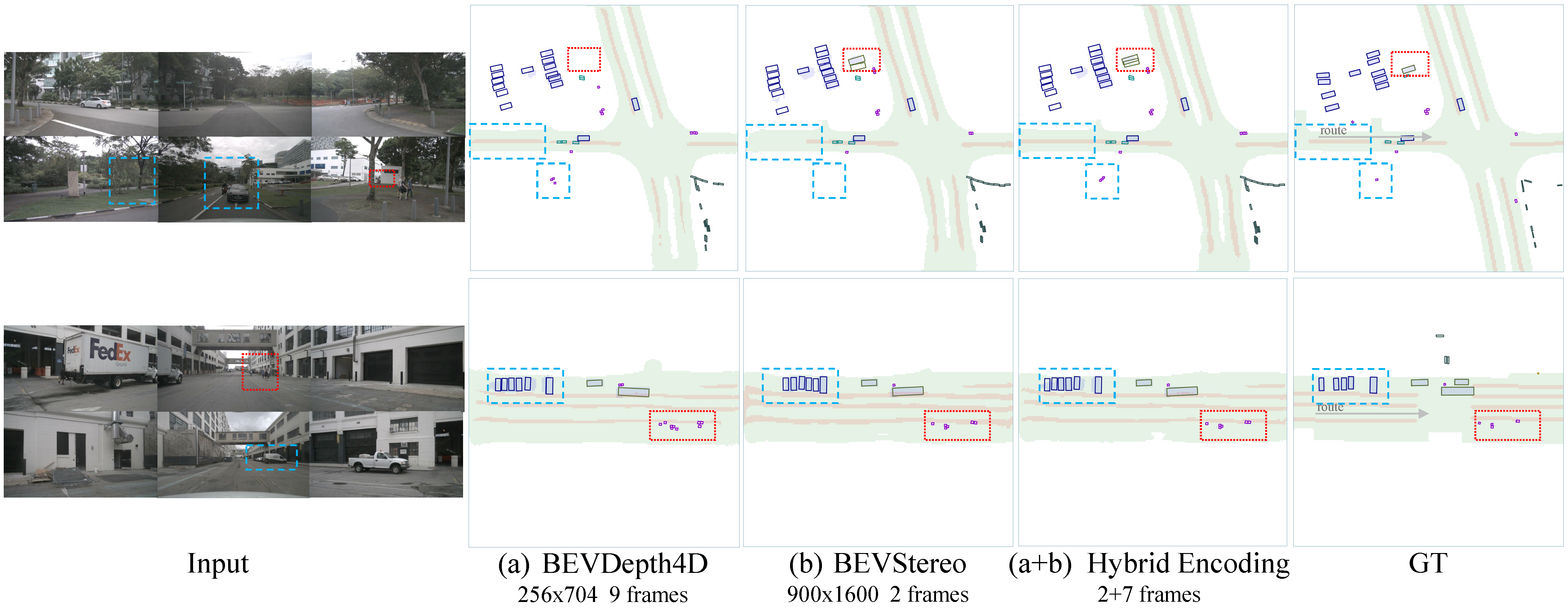}
 \caption{\textbf{Visualization results of Hybrid Encoding and baseline methods.}
From left to right: multi-view image inputs, BEVDepth4D, BEVStereo, Hybrid Encoding results, and ground truth.
Hybrid Encoding improves occluded object estimation by leveraging long-term temporal information and yields more accurate predictions using high-resolution features.
}
    \label{fig:vis}

\end{figure*}

\subsection{Single-Task Results}

Since many prior works in 3D perception report only single-task performance, we evaluate HENet++ under single-task settings and compare with task-specific models.
These comparisons highlight the effectiveness of the proposed Hybrid Image Encoding Network and Temporal Feature Integration, and further validate the design of HENet++.

\vspace{1mm}
\noindent \textbf{3D Object Detection Results.}
Single-task 3D object detection results on the nuScenes \texttt{val} and \texttt{test} sets are reported in~\tabref{table:exp2} and~\tabref{table:exp3}, respectively.
As shown in~\tabref{table:exp2}, HENet and HENet++ outperform existing multi-view camera-based detection methods across different backbone configurations, demonstrating the effectiveness of hybrid image encoding and temporal integration.
Results in~\tabref{table:exp3} further show that HENet++ achieves state-of-the-art performance.
By transitioning from BEV-centric representations to sparse instance modeling, HENet++ improves detection accuracy over HENet.
Compared with its primary baseline SparseBEV, the gains also validate the effectiveness of Hybrid Encoding.

\vspace{1mm}
\noindent \textbf{BEV Semantic Segmentation Results.}
Results on the nuScenes \texttt{val} set are shown in~\tabref{table:exp4}.
HENet++ achieves competitive performance compared with existing methods.
Since BEV semantic segmentation is decoded from dense features and the single-task architecture remains unchanged, HENet and HENet++ share identical results and are reported in the same row.

\vspace{1mm}
\noindent \textbf{Occupancy Results.}
Single-task occupancy results on the nuScenes \texttt{val} set are presented in Table 6, where HENet++ achieves state-of-the-art performance.
Compared with its primary baseline FB-Occ, the gains also validate the effectiveness of Hybrid Encoding.
We further report RayIoU alongside mIoU to provide a complementary evaluation that is less sensitive to thickness and visibility inflation.

Notably, compared with single-task training, the end-to-end multi-task setting introduces only minor performance drops: 0.4 mAP for 3D object detection, 0.5 mIoU for BEV semantic segmentation, and 0.8 mIoU for occupancy prediction, demonstrating strong task compatibility.






\setlength{\tabcolsep}{0.4em}
\begin{table}[t]
\centering
\tablestyle{6pt}{1.2}
\caption{\textbf{Ablation of Pretrain Method of HENet++.}
This experiment is conducted on a triple-task R50 small model, repeated five times with different random seeds, and reports the mean and standard deviation.
Compared to using single-task detection or occupancy weights, loading the merged model for pre-training improves multi-task performance. 
}

\label{table:abl4} 
\begin{tabular}{l|ccc}
\toprule
Load from & \textbf{NDS} & $\textbf{mIoU}_{bev}$ & $\textbf{mIoU}_{occ}$ \\

\hline
Detection & 53.6 $\pm$ 0.2 & 54.3 $\pm$ 0.5 & 36.2 $\pm$ 0.3 \\
\gr Segmentation & 52.8 $\pm$ 0.4 & 55.0 $\pm$ 0.3 & 34.9 $\pm$ 0.6 \\
Occupancy & 53.2 $\pm$ 0.2 & 54.9 $\pm$ 0.5 & 37.3 $\pm$ 0.2 \\
\hepp Model Merge & \textbf{53.8} $\pm$ 0.3 & \textbf{55.3} $\pm$ 0.3 & \textbf{37.6} $\pm$ 0.2 \\

\bottomrule
\end{tabular}
\end{table}

\setlength{\tabcolsep}{0.4em}
\begin{table}[t]
\setlength{\belowcaptionskip}{10pt}
\centering
\tablestyle{1.5pt}{1.2}
\caption{\textbf{Ablation on the Loss for End-to-End Autonomous Driving.}
}

\label{table:abl5} 
\begin{tabular}{ccc|cccc}
\toprule
\multirow{2}{*}{$\mathcal{L}_{plan}$} & \multirow{2}{*}{$\mathcal{L}_{pred}$} & \multirow{2}{*}{$\mathcal{L}_{col}$} & \multicolumn{2}{c}{UniAD Metrics} & \multicolumn{2}{c}{VAD/STP3 Metrics} \\
& & & mean L2 & mean Col. & mean L2 & mean Col. \\

\hline
\checkmark & & & 1.69 & 0.52 & 0.71 & 0.17 \\
\gr \checkmark & \checkmark & & 1.36 & 0.20 & 0.58 & 0.10 \\
\hepp \checkmark & \checkmark & \checkmark & 1.29 & 0.13 & 0.55 & 0.05 \\

\bottomrule
\end{tabular}
\end{table}

\begin{figure}[t]
    \setlength{\abovecaptionskip}{10pt}
    \centering
    \includegraphics[width=\linewidth]{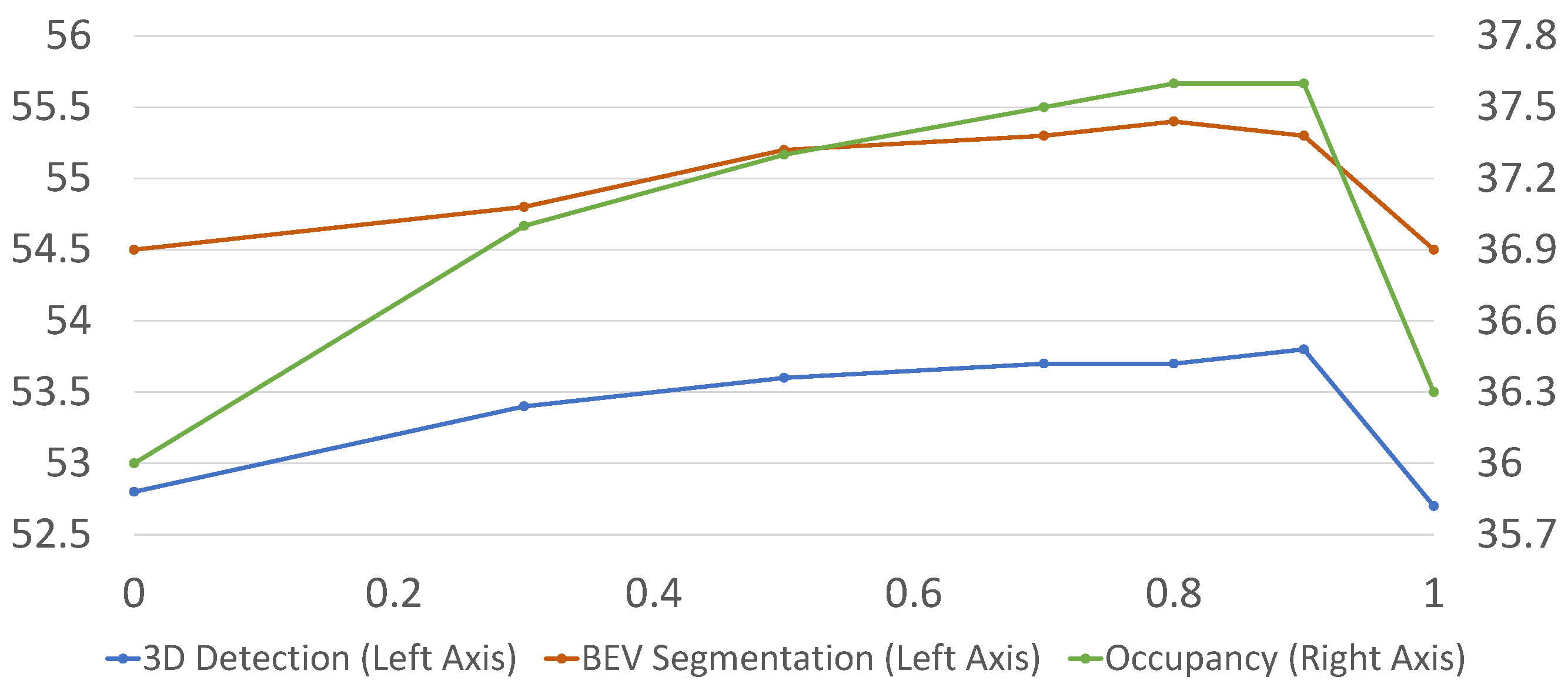}
    \caption{\textbf{Ablation study on the hyperparameter $\alpha$ for model merging.}}
    \label{fig:alpha}
\end{figure}

\subsection{Ablation Studies}

We conduct ablation studies on the nuScenes \texttt{val} set to evaluate the contribution of each proposed component.

\vspace{1mm}
\noindent \textbf{Hybrid Image Encoding Network.}
To validate the effectiveness of the proposed hybrid image encoding network, we compare HENet++ with three baseline models and their ensemble variants.
As shown in \tabref{table:abl1}, combining BEVDepth4D~\citep{BEVDepth} and BEVStereo~\citep{BEVStereo} through hybrid image encoding, together with the AFFM module, significantly improves 3D object detection performance.
Compared with increasing image resolution alone (model C), the Hybrid Image Encoding Network achieves higher accuracy, faster inference speed, and lower training cost.
Similarly, compared with increasing the temporal length (model F), our approach achieves higher accuracy while reducing training cost.

Notably, simple ensembling of the two baselines with NMS degrades performance as the weaker BEVDepth4D~\citep{BEVDepth} introduces additional false positives.
Consistent improvements are also observed when comparing SparseBEV~\citep{SparseBEV} with HENet++, demonstrating that Hybrid Encoding benefits both BEV-based and sparse-feature-based perception frameworks.

For comprehensive evaluation, multi-task ablation results are presented in \tabref{table:abl1b}, and qualitative visualizations are shown in \figref{fig:vis}.
As illustrated, certain challenging scenarios involving motion or occlusion (blue boxes) benefit from longer temporal context, while high-resolution inputs and more accurate depth estimation improve perception in complex scenes (red boxes).
Hybrid Encoding effectively combines long-term temporal information, high-resolution features, and advanced depth estimation to achieve improved performance.





\vspace{1mm}
\noindent \textbf{Pre-training via Model Merging.}
As shown in~\tabref{table:abl4}, initializing the multi-task model with merged weights from multiple single-task models consistently improves performance compared to using weights from individual detection or occupancy models alone.
These results suggest that model merging provides a more balanced initialization, mitigating task bias introduced by single-task pre-training and facilitating better joint optimization across tasks.

\vspace{1mm}
\noindent \textbf{Loss Design for End-to-End Autonomous Driving.}
Experimental results in~\tabref{table:abl5} demonstrate the effectiveness of the proposed loss formulation described in~\secref{sec:e2e}.
In particular, combining planning, prediction, and collision-aware objectives improves trajectory quality and reduces unsafe behaviors, indicating that the joint loss design provides complementary supervision for stable end-to-end learning.

\vspace{1mm}
\noindent \textbf{The hyperparameter $\alpha$ in model merging.}
The ablation study on the parameter $\alpha$ is shown in~\figref{fig:alpha}. Model merging is not particularly sensitive to the choice of $\alpha$. Performance across values ranging from 0.7 to 0.9 is comparable and falls within the standard deviation range.


\section{Conclusion}

In this paper, we introduce HENet++, a unified framework for end-to-end multi-task 3D perception.
By jointly encoding sparse foreground representations and dense background voxel features through hybrid encoding, HENet++ provides task-suitable representations for 3D object detection, BEV semantic segmentation, and occupancy prediction within a single model.
We further propose a model-merging-based pre-training strategy that improves multi-task optimization and overall performance.
Experimental results demonstrate that HENet++ achieves state-of-the-art end-to-end multi-task perception performance on the nuScenes dataset.

Building upon this perception framework, we develop an end-to-end autonomous driving model that leverages complementary sparse and dense representations.
An attention-based world-prediction module enables simultaneous trajectory prediction and ego-vehicle planning.
HENet++ achieves a lower collision rate than existing approaches on nuScenes and achieves higher PDMS on the NAVSIM benchmark.


\backmatter

\section*{Declarations}

\bmhead{Preliminary Version}

Preliminary results of this work were presented in HENet~\citep{xia2024henet}.

\bmhead{Acknowledgements}

This work was supported by the Beijing Municipal Natural Science Foundation (Grant No. L2603008), the National Key R\&D Program of China (Grant No. 2022ZD0160305), and the National Natural Science Foundation of China (Grant No. 62176007).

\bmhead{Data Availability}

All experiments were conducted using publicly available datasets.
The nuScenes dataset is available at \url{https://www.nuscenes.org/nuscenes}, and the nuScenes Occupancy Ground Truth is available at \url{https://github.com/Tsinghua-MARS-Lab/Occ3D}.
The NAVSIM benchmark is available at \url{https://github.com/autonomousvision/navsim/tree/v1.1}.

\bibliography{ref_main}

\begin{thebibliography}{76}
\providecommand{\natexlab}[1]{#1}
\providecommand{\url}[1]{{#1}}
\providecommand{\urlprefix}{URL }
\providecommand{\doi}[1]{\url{https://doi.org/#1}}
\providecommand{\eprint}[2][]{\url{#2}}
 \bibcommenthead

\bibitem[{Badrinarayanan et~al.(2017)Badrinarayanan, Kendall, and Cipolla}]{segnet}
Badrinarayanan V, Kendall A, Cipolla R (2017) Segnet: A deep convolutional encoder-decoder architecture for image segmentation. IEEE Transactions on Pattern Recognition and Machine Intelligence (TPAMI)

\bibitem[{Brazil and Liu(2019)}]{M3d-rpn}
Brazil G, Liu X (2019) M3d-rpn: Monocular 3d region proposal network for object detection. In: International Conference on Computer Vision (ICCV)

\bibitem[{Caesar et~al.(2020)Caesar, Bankiti, Lang, Vora, Liong, Xu, Krishnan, Pan, Baldan, and Beijbom}]{nuScenes}
Caesar H, Bankiti V, Lang AH, et~al (2020) nuscenes: A multimodal dataset for autonomous driving. In: IEEE/CVF Conference on Computer Vision and Pattern Recognition (CVPR)

\bibitem[{Cao and De~Charette(2022)}]{22}
Cao AQ, De~Charette R (2022) Monoscene: Monocular 3d semantic scene completion. In: IEEE/CVF Conference on Computer Vision and Pattern Recognition (CVPR)

\bibitem[{Casas et~al.(2021)Casas, Sadat, and Urtasun}]{mp3}
Casas S, Sadat A, Urtasun R (2021) Mp3: A unified model to map, perceive, predict and plan. In: IEEE/CVF Conference on Computer Vision and Pattern Recognition (CVPR)

\bibitem[{Dauner et~al.(2024)Dauner, Hallgarten, Li, Weng, Huang, Yang, Li, Gilitschenski, Ivanovic, Pavone et~al.}]{dauner2024navsim}
Dauner D, Hallgarten M, Li T, et~al (2024) Navsim: Data-driven non-reactive autonomous vehicle simulation and benchmarking. In: Conference on Neural Information Processing Systems (NeurIPS)

\bibitem[{Ding et~al.(2020)Ding, Huo, Yi, Wang, Shi, Lu, and Luo}]{D4LCN}
Ding M, Huo Y, Yi H, et~al (2020) Learning depth-guided convolutions for monocular 3d object detection. In: IEEE/CVF Conference on Computer Vision and Pattern Recognition (CVPR)

\bibitem[{Feng et~al.(2023)Feng, Jie, Zhong, Chu, and Ma}]{aedet}
Feng C, Jie Z, Zhong Y, et~al (2023) Aedet: Azimuth-invariant multi-view 3d object detection. In: IEEE/CVF Conference on Computer Vision and Pattern Recognition (CVPR)

\bibitem[{He et~al.(2016)He, Zhang, Ren, and Sun}]{resnet}
He K, Zhang X, Ren S, et~al (2016) Deep residual learning for image recognition. In: IEEE/CVF Conference on Computer Vision and Pattern Recognition (CVPR)

\bibitem[{Hou et~al.(2024)Hou, Li, Guan, Zhang, Feng, Du, Xue, and Pu}]{31}
Hou J, Li X, Guan W, et~al (2024) Fastocc: Accelerating 3d occupancy prediction by fusing the 2d bird's-eye view and perspective view. In: IEEE International Conference on Robotics and Automation (ICRA)

\bibitem[{Hu et~al.(2021)Hu, Murez, Mohan, Dudas, Hawke, Badrinarayanan, Cipolla, and Kendall}]{fiery}
Hu A, Murez Z, Mohan N, et~al (2021) Fiery: Future instance prediction in bird's-eye view from surround monocular cameras. In: International Conference on Computer Vision (ICCV)

\bibitem[{Hu et~al.(2022)Hu, Chen, Wu, Li, Yan, and Tao}]{hu2022st}
Hu S, Chen L, Wu P, et~al (2022) St-p3: End-to-end vision-based autonomous driving via spatial-temporal feature learning. In: European Conference on Computer Vision (ECCV)

\bibitem[{Hu et~al.(2023)Hu, Yang, Chen, Li, Sima, Zhu, Chai, Du, Lin, Wang, Lu, Jia, Liu, Dai, Qiao, and Li}]{UniAD}
Hu Y, Yang J, Chen L, et~al (2023) Planning-oriented autonomous driving. In: IEEE/CVF Conference on Computer Vision and Pattern Recognition (CVPR)

\bibitem[{Huang et~al.(2023{\natexlab{a}})Huang, Li, Xie, Liang, Wang, Shen, Liu, Wang, Luo, and Shao}]{fastBEV}
Huang B, Li Y, Xie E, et~al (2023{\natexlab{a}}) Fast-bev: Towards real-time on-vehicle bird's-eye view perception. arXiv preprint arXiv:230107870

\bibitem[{Huang and Huang(2022)}]{bevpoolv2}
Huang J, Huang G (2022) Bevpoolv2: A cutting-edge implementation of bevdet toward deployment. arXiv preprint arXiv:221117111

\bibitem[{Huang et~al.(2021)Huang, Huang, Zhu, Ye, and Du}]{BEVDet}
Huang J, Huang G, Zhu Z, et~al (2021) Bevdet: High-performance multi-camera 3d object detection in bird-eye-view. arXiv preprint arXiv:211211790

\bibitem[{Huang et~al.(2022)Huang, Huang, and Robotics}]{BEVDet4D}
Huang J, Huang G, Robotics P (2022) Bevdet4d: Exploit temporal cues in multi-camera 3d object detection. arXiv preprint arXiv:220317054

\bibitem[{Huang et~al.(2023{\natexlab{b}})Huang, Zheng, Zhang, Zhou, and Lu}]{23}
Huang Y, Zheng W, Zhang Y, et~al (2023{\natexlab{b}}) Tri-perspective view for vision-based 3d semantic occupancy prediction. In: IEEE/CVF Conference on Computer Vision and Pattern Recognition (CVPR)

\bibitem[{Javaloy and Valera(2021)}]{javaloy2021rotograd}
Javaloy A, Valera I (2021) Rotograd: Gradient homogenization in multitask learning. arXiv preprint arXiv:210302631

\bibitem[{Jiang et~al.(2023{\natexlab{a}})Jiang, Chen, Xu, Liao, Chen, Zhou, Zhang, Liu, Huang, and Wang}]{jiang2023vad}
Jiang B, Chen S, Xu Q, et~al (2023{\natexlab{a}}) Vad: Vectorized scene representation for efficient autonomous driving. In: International Conference on Computer Vision (ICCV)

\bibitem[{Jiang et~al.(2023{\natexlab{b}})Jiang, Chen, Pan, Wang, Liu, Jiang, and Long}]{jiang2023forkmerge}
Jiang J, Chen B, Pan J, et~al (2023{\natexlab{b}}) Forkmerge: Mitigating negative transfer in auxiliary-task learning. In: Conference on Neural Information Processing Systems (NeurIPS)

\bibitem[{Jiang et~al.(2024)Jiang, Li, Liu, Wang, Jia, Wang, Han, and Zhang}]{far3d}
Jiang X, Li S, Liu Y, et~al (2024) Far3d: Expanding the horizon for surround-view 3d object detection. In: Association for the Advancement of Artificial Intelligence (AAAI)

\bibitem[{Jiang et~al.(2023{\natexlab{c}})Jiang, Zhang, Miao, Zhu, Gao, Hu, and Jiang}]{polarformer}
Jiang Y, Zhang L, Miao Z, et~al (2023{\natexlab{c}}) Polarformer: Multi-camera 3d object detection with polar transformer. In: Association for the Advancement of Artificial Intelligence (AAAI)

\bibitem[{Jin et~al.(2023)Jin, Ren, Preotiuc-Pietro, and Cheng}]{regmean}
Jin X, Ren X, Preotiuc-Pietro D, et~al (2023) Dataless knowledge fusion by merging weights of language models. In: International Conference on Learning Representations (ICLR)

\bibitem[{Lee and Park(2020)}]{v299}
Lee Y, Park J (2020) Centermask: Real-time anchor-free instance segmentation. In: IEEE/CVF Conference on Computer Vision and Pattern Recognition (CVPR)

\bibitem[{Li et~al.(2022{\natexlab{a}})Li, Wang, Wang, and Zhao}]{li2022hdmapnet}
Li Q, Wang Y, Wang Y, et~al (2022{\natexlab{a}}) Hdmapnet: An online hd map construction and evaluation framework. In: IEEE International Conference on Robotics and Automation (ICRA)

\bibitem[{Li et~al.(2023{\natexlab{a}})Li, Bao, Ge, Yang, Sun, and Li}]{BEVStereo}
Li Y, Bao H, Ge Z, et~al (2023{\natexlab{a}}) Bevstereo: Enhancing depth estimation in multi-view 3d object detection with dynamic temporal stereo. In: Association for the Advancement of Artificial Intelligence (AAAI)

\bibitem[{Li et~al.(2023{\natexlab{b}})Li, Ge, Yu, Yang, Wang, Shi, Sun, and Li}]{BEVDepth}
Li Y, Ge Z, Yu G, et~al (2023{\natexlab{b}}) Bevdepth: Acquisition of reliable depth for multi-view 3d object detection. In: Association for the Advancement of Artificial Intelligence (AAAI)

\bibitem[{Li et~al.(2023{\natexlab{c}})Li, Yu, Choy, Xiao, Alvarez, Fidler, Feng, and Anandkumar}]{25}
Li Y, Yu Z, Choy C, et~al (2023{\natexlab{c}}) Voxformer: Sparse voxel transformer for camera-based 3d semantic scene completion. In: IEEE/CVF Conference on Computer Vision and Pattern Recognition (CVPR)

\bibitem[{Li et~al.(2022{\natexlab{b}})Li, Wang, Li, Xie, Sima, Lu, Qiao, and Dai}]{BEVFormer}
Li Z, Wang W, Li H, et~al (2022{\natexlab{b}}) Bevformer: Learning bird’s-eye-view representation from multi-camera images via spatiotemporal transformers. In: European Conference on Computer Vision (ECCV)

\bibitem[{Li et~al.(2023{\natexlab{d}})Li, Yu, Austin, Fang, Lan, Kautz, and Alvarez}]{27}
Li Z, Yu Z, Austin D, et~al (2023{\natexlab{d}}) Fb-occ: 3d occupancy prediction based on forward-backward view transformation. arXiv preprint arXiv:230701492

\bibitem[{Liao et~al.(2025)Liao, Chen, Yin, Jiang, Wang, Yan, Zhang, Li, Zhang, Zhang et~al.}]{liao2025diffusiondrive}
Liao B, Chen S, Yin H, et~al (2025) Diffusiondrive: Truncated diffusion model for end-to-end autonomous driving. In: IEEE/CVF Conference on Computer Vision and Pattern Recognition (CVPR)

\bibitem[{Lin et~al.(2017)Lin, Doll{\'a}r, Girshick, He, Hariharan, and Belongie}]{FPN}
Lin TY, Doll{\'a}r P, Girshick R, et~al (2017) Feature pyramid networks for object detection. In: IEEE/CVF Conference on Computer Vision and Pattern Recognition (CVPR)

\bibitem[{Lin et~al.(2022)Lin, Lin, Pei, Huang, and Su}]{sparse4d}
Lin X, Lin T, Pei Z, et~al (2022) Sparse4d: Multi-view 3d object detection with sparse spatial-temporal fusion. arXiv preprint arXiv:221110581

\bibitem[{Lin et~al.(2023)Lin, Lin, Pei, Huang, and Su}]{sparse4dv2}
Lin X, Lin T, Pei Z, et~al (2023) Sparse4d v2: Recurrent temporal fusion with sparse model. arXiv preprint arXiv:230514018

\bibitem[{Lin et~al.(2024)Lin, Liu, Wang, Zhang, and Zhu}]{lin2024rcbevdet++}
Lin Z, Liu Z, Wang Y, et~al (2024) Rcbevdet++: toward high-accuracy radar-camera fusion 3d perception network. arXiv preprint arXiv:240904979

\bibitem[{Liu et~al.(2023{\natexlab{a}})Liu, Teng, Lu, Wang, and Wang}]{SparseBEV}
Liu H, Teng Y, Lu T, et~al (2023{\natexlab{a}}) Sparsebev: High-performance sparse 3d object detection from multi-camera videos. In: International Conference on Computer Vision (ICCV)

\bibitem[{Liu et~al.(2025)Liu, Wang, Qiao, Guo, Zhu, Li, Chen, Yang, Li, Wang et~al.}]{liu2025mmtl}
Liu W, Wang W, Qiao Y, et~al (2025) Mmtl-uniad: A unified framework for multimodal and multi-task learning in assistive driving perception. In: IEEE/CVF Conference on Computer Vision and Pattern Recognition (CVPR)

\bibitem[{Liu et~al.(2022)Liu, Wang, Zhang, and Sun}]{PETR}
Liu Y, Wang T, Zhang X, et~al (2022) Petr: Position embedding transformation for multi-view 3d object detection. In: European Conference on Computer Vision (ECCV)

\bibitem[{Liu et~al.(2023{\natexlab{b}})Liu, Yan, Jia, Li, Gao, Wang, and Zhang}]{PETRv2}
Liu Y, Yan J, Jia F, et~al (2023{\natexlab{b}}) Petrv2: A unified framework for 3d perception from multi-camera images. In: International Conference on Computer Vision (ICCV)

\bibitem[{Ma et~al.(2024)Ma, Tan, Qu, Ma, Zhang, and Xie}]{32}
Ma Q, Tan X, Qu Y, et~al (2024) Cotr: Compact occupancy transformer for vision-based 3d occupancy prediction. In: IEEE/CVF Conference on Computer Vision and Pattern Recognition (CVPR)

\bibitem[{Pan et~al.(2020)Pan, Sun, Leung, Andonian, and Zhou}]{VPN}
Pan B, Sun J, Leung HYT, et~al (2020) Cross-view semantic segmentation for sensing surroundings. IEEE Robotics and Automation Letters (RAL)

\bibitem[{Pan et~al.(2023)Pan, Liu, Liu, Huang, Wang, Zhang, Xu, Lai, and Yang}]{28}
Pan M, Liu L, Liu J, et~al (2023) Uniocc: Unifying vision-centric 3d occupancy prediction with geometric and semantic rendering. arXiv preprint arXiv:230609117

\bibitem[{Pan et~al.(2024)Pan, Liu, Zhang, Huang, Li, Liu, and Zhang}]{29}
Pan M, Liu J, Zhang R, et~al (2024) Renderocc: Vision-centric 3d occupancy prediction with 2d rendering supervision. In: IEEE International Conference on Robotics and Automation (ICRA)

\bibitem[{Park et~al.(2021)Park, Ambrus, Guizilini, Li, and Gaidon}]{dd3d}
Park D, Ambrus R, Guizilini V, et~al (2021) Is pseudo-lidar needed for monocular 3d object detection? In: International Conference on Computer Vision (ICCV)

\bibitem[{Park et~al.(2023)Park, Xu, Yang, Keutzer, Kitani, Tomizuka, and Zhan}]{SOLOFusion}
Park J, Xu C, Yang S, et~al (2023) Time will tell: New outlooks and a baseline for temporal multi-view 3d object detection. In: International Conference on Learning Representations (ICLR)

\bibitem[{Patapati(2025)}]{patapati2025graph}
Patapati S (2025) Graph coloring for multi-task learning. arXiv preprint arXiv:250916959

\bibitem[{Philion and Fidler(2020)}]{LSS}
Philion J, Fidler S (2020) Lift, splat, shoot: Encoding images from arbitrary camera rigs by implicitly unprojecting to 3d. In: European Conference on Computer Vision (ECCV)

\bibitem[{Reading et~al.(2021)Reading, Harakeh, Chae, and Waslander}]{CaDNN}
Reading C, Harakeh A, Chae J, et~al (2021) Categorical depth distribution network for monocular 3d object detection. In: IEEE/CVF Conference on Computer Vision and Pattern Recognition (CVPR)

\bibitem[{Roddick and Cipolla(2020)}]{roddick2020predicting}
Roddick T, Cipolla R (2020) Predicting semantic map representations from images using pyramid occupancy networks. In: IEEE/CVF Conference on Computer Vision and Pattern Recognition (CVPR)

\bibitem[{Roddick et~al.(2018)Roddick, Kendall, and Cipolla}]{OFT}
Roddick T, Kendall A, Cipolla R (2018) Orthographic feature transform for monocular 3d object detection. arXiv preprint arXiv:181108188

\bibitem[{Song et~al.(2025)Song, Jia, Liu, Pan, Zhang, Wang, Zhang, Xu, Yang, and Luo}]{song2025don}
Song Z, Jia C, Liu L, et~al (2025) Don't shake the wheel: Momentum-aware planning in end-to-end autonomous driving. In: IEEE/CVF Conference on Computer Vision and Pattern Recognition (CVPR)

\bibitem[{Sun et~al.(2025)Sun, Lin, Shi, Zhang, Wu, and Zheng}]{sun2024sparsedrive}
Sun W, Lin X, Shi Y, et~al (2025) Sparsedrive: End-to-end autonomous driving via sparse scene representation. In: IEEE International Conference on Robotics and Automation (ICRA)

\bibitem[{Tang et~al.(2024)Tang, Wang, Wang, Zheng, Ren, Feng, and Ma}]{tang2024sparseocc}
Tang P, Wang Z, Wang G, et~al (2024) Sparseocc: Rethinking sparse latent representation for vision-based semantic occupancy prediction. In: IEEE/CVF Conference on Computer Vision and Pattern Recognition (CVPR)

\bibitem[{Tong et~al.(2023)Tong, Sima, Wang, Chen, Wu, Deng, Gu, Lu, Luo, Lin et~al.}]{occnet}
Tong W, Sima C, Wang T, et~al (2023) Scene as occupancy. In: International Conference on Computer Vision (ICCV)

\bibitem[{Wang et~al.(2023)Wang, Liu, Wang, Li, and Zhang}]{StreamPetr}
Wang S, Liu Y, Wang T, et~al (2023) Exploring object-centric temporal modeling for efficient multi-view 3d object detection. In: International Conference on Computer Vision (ICCV)

\bibitem[{Wang et~al.(2021{\natexlab{a}})Wang, Zhu, Pang, and Lin}]{Fcos3d}
Wang T, Zhu X, Pang J, et~al (2021{\natexlab{a}}) Fcos3d: Fully convolutional one-stage monocular 3d object detection. In: International Conference on Computer Vision (ICCV)

\bibitem[{Wang et~al.(2022{\natexlab{a}})Wang, Pang, and Lin}]{DFM}
Wang T, Pang J, Lin D (2022{\natexlab{a}}) Monocular 3d object detection with depth from motion. In: European Conference on Computer Vision (ECCV)

\bibitem[{Wang et~al.(2019)Wang, Chao, Garg, Hariharan, Campbell, and Weinberger}]{wang2019pseudo}
Wang Y, Chao WL, Garg D, et~al (2019) Pseudo-lidar from visual depth estimation: Bridging the gap in 3d object detection for autonomous driving. In: IEEE/CVF Conference on Computer Vision and Pattern Recognition (CVPR)

\bibitem[{Wang et~al.(2021{\natexlab{b}})Wang, Guizilini, Zhang, Wang, Zhao, and Solomon}]{DETR3D}
Wang Y, Guizilini V, Zhang T, et~al (2021{\natexlab{b}}) Detr3d: 3d object detection from multi-view images via 3d-to-2d queries. In: Conference on Robot Learning (CoRL)

\bibitem[{Wang et~al.(2022{\natexlab{b}})Wang, Min, Ge, Li, Li, Yang, and Huang}]{sts}
Wang Z, Min C, Ge Z, et~al (2022{\natexlab{b}}) Sts: Surround-view temporal stereo for multi-view 3d detection. arXiv preprint arXiv:220810145

\bibitem[{Wei et~al.(2023)Wei, Zhao, Zheng, Zhu, Zhou, and Lu}]{24}
Wei Y, Zhao L, Zheng W, et~al (2023) Surroundocc: Multi-camera 3d occupancy prediction for autonomous driving. In: International Conference on Computer Vision (ICCV)

\bibitem[{Weng et~al.(2024)Weng, Ivanovic, Wang, Wang, and Pavone}]{weng2024drive}
Weng X, Ivanovic B, Wang Y, et~al (2024) Para-drive: Parallelized architecture for real-time autonomous driving. In: IEEE/CVF Conference on Computer Vision and Pattern Recognition (CVPR)

\bibitem[{Xia et~al.(2024)Xia, Lin, Wang, Wang, Xing, Qi, Dong, and Yang}]{xia2024henet}
Xia Z, Lin Z, Wang X, et~al (2024) Henet: Hybrid encoding for end-to-end multi-task 3d perception from multi-view cameras. In: European Conference on Computer Vision (ECCV)

\bibitem[{Xie et~al.(2022)Xie, Yu, Zhou, Philion, Anandkumar, Fidler, Luo, and Alvarez}]{xie2022m}
Xie E, Yu Z, Zhou D, et~al (2022) M2bev: Multi-camera joint 3d detection and segmentation with unified birds-eye view representation. arXiv preprint arXiv:220405088

\bibitem[{Yang et~al.(2023)Yang, Chen, Tian, Tao, Zhu, Zhang, Huang, Li, Qiao, Lu et~al.}]{bevformerv2}
Yang C, Chen Y, Tian H, et~al (2023) Bevformer v2: Adapting modern image backbones to bird's-eye-view recognition via perspective supervision. In: IEEE/CVF Conference on Computer Vision and Pattern Recognition (CVPR)

\bibitem[{Yang et~al.(2021{\natexlab{a}})Yang, Li, Liu, Yu, Ma, He, and Pan}]{yang2021projecting}
Yang W, Li Q, Liu W, et~al (2021{\natexlab{a}}) Projecting your view attentively: Monocular road scene layout estimation via cross-view transformation. In: IEEE/CVF Conference on Computer Vision and Pattern Recognition (CVPR)

\bibitem[{Yang et~al.(2021{\natexlab{b}})Yang, Zhou, Chen, and Ngiam}]{3d-man}
Yang Z, Zhou Y, Chen Z, et~al (2021{\natexlab{b}}) 3d-man: 3d multi-frame attention network for object detection. In: IEEE/CVF Conference on Computer Vision and Pattern Recognition (CVPR)

\bibitem[{Yu et~al.(2023)Yu, Shu, Deng, Lu, Liu, Yu, Yang, Li, and Chen}]{26}
Yu Z, Shu C, Deng J, et~al (2023) Flashocc: Fast and memory-efficient occupancy prediction via channel-to-height plugin. arXiv preprint arXiv:231112058

\bibitem[{Yuan et~al.(2018)Yuan, Huang, Guo, Zhang, Chen, and Wang}]{OCNet}
Yuan Y, Huang L, Guo J, et~al (2018) Ocnet: Object context network for scene parsing. arXiv preprint arXiv:180900916

\bibitem[{Zhang et~al.(2025)Zhang, Song, Jin, and Zhang}]{zhang2025bridging}
Zhang B, Song N, Jin X, et~al (2025) Bridging past and future: End-to-end autonomous driving with historical prediction and planning. In: IEEE/CVF Conference on Computer Vision and Pattern Recognition (CVPR)

\bibitem[{Zhang et~al.(2023)Zhang, Zheng, Zhu, Huang, Lu, and Zhou}]{simMOD}
Zhang Y, Zheng W, Zhu Z, et~al (2023) A simple baseline for multi-camera 3d object detection. In: Association for the Advancement of Artificial Intelligence (AAAI)

\bibitem[{Zheng et~al.(2024)Zheng, Song, Guo, and Chen}]{zheng2024genad}
Zheng W, Song R, Guo X, et~al (2024) Genad: Generative end-to-end autonomous driving. In: European Conference on Computer Vision (ECCV)

\bibitem[{Zhou and Kr{\"a}henb{\"u}hl(2022)}]{zhou2022cross}
Zhou B, Kr{\"a}henb{\"u}hl P (2022) Cross-view transformers for real-time map-view semantic segmentation. In: IEEE/CVF Conference on Computer Vision and Pattern Recognition (CVPR)

\bibitem[{Zhuang et~al.(2021)Zhuang, Li, Jia, Wang, Li, and Tan}]{PMF}
Zhuang Z, Li R, Jia K, et~al (2021) Perception-aware multi-sensor fusion for 3d lidar semantic segmentation. In: International Conference on Computer Vision (ICCV)

\bibitem[{Zong et~al.(2023)Zong, Jiang, Song, Xue, Su, Li, and Liu}]{HoP}
Zong Z, Jiang D, Song G, et~al (2023) Temporal enhanced training of multi-view 3d object detector via historical object prediction. In: International Conference on Computer Vision (ICCV)

\end{thebibliography}

\end{document}